%% file: main.tex
\pdfoutput=1

\documentclass[11pt]{article}
\usepackage{arabtex}
\usepackage[preprint]{acl}
\usepackage{enumitem}

\usepackage{times}
\usepackage{latexsym}
\usepackage{amsmath}

\usepackage[T1]{fontenc}
\usepackage[utf8]{inputenc}
\usepackage{todonotes}

\usepackage{microtype}
\usepackage{inconsolata}

\usepackage{graphicx}
\usepackage{booktabs}

\usepackage{contour}

\input{macros}

\usepackage{utf8}
\usepackage{epstopdf}

\newcommand{\cmtb}{{\sc MadarTB}}
\newcommand{\pos}{{\sc POS}}

\title{A Parallel Multi-Dialect Arabic Dependency Treebank}
\author{
  Noor Abo Mokh\textsuperscript{2}\thanks{Equal contribution.}\quad 
  Mohamed Adel\textsuperscript{1}\footnotemark[1]\quad 
  Jamila El Gizuli\textsuperscript{1}\\
  \textbf{Go Inoue\textsuperscript{1,2}\quad 
  Ahmed Elshabrawy\textsuperscript{1,2}\quad 
  Nizar Habash\textsuperscript{1}}
  \\
  \textsuperscript{1}Computational Approaches to Modeling Language Lab, New York University Abu Dhabi\\
  \textsuperscript{2}Mohamed bin Zayed University of Artificial Intelligence\\
  \texttt{\{noor.abomokh, go.inoue, ahmed.elshabrawy\}@mbzuai.ac.ae}\\
  \texttt{\{mohamed.adel, nizar.habash\}@nyu.edu}
}

\begin{document}
\maketitle

\setcode{utf8}
\vocalize

\begin{abstract}
Dependency parsing resource development for Arabic has largely focused on Modern Standard Arabic, leaving dialectal Arabic comparatively under-resourced. We introduce {\cmtb}, the first large-scale parallel dependency treebank spanning five city dialects: Beirut, Cairo, Doha, Rabat and Tunis, and extending an existing parallel MSA version. In total {\cmtb} consists of 5,000 sentences annotated using the CATiB dependency scheme. 
We analyze syntactic similarity across all city dialects and MSA. Although global construction profiles exhibit high similarity across varieties, local syntactic structures frequently diverge, revealing substantial structural variation across dialects.
We further establish a cross-dialect dependency parsing comparison using a supervised parser and four LLMs. The supervised parser achieves 87.5\% LAS, while the best LLM setup trails at 85.7\% LAS across all six varieties on the test set.
We release our annotations and models publicly.

\end{abstract}

\section{Introduction}
Dependency parsing represents grammatical structure as labeled relations between words and supports both linguistic analysis and NLP applications \citep{Jurafsky:2009:SLP:1214993}. Beyond probing whether LLMs capture grammatical relations \citep{kulmizev2023search}, it enables syntax-aware applications such as information extraction, error detection, writing feedback, and grammatical error correction \citep{kanayama-etal-2024,li-etal-2025-generation,metheniti-etal-2019-identifying,morgado-da-costa-etal-2020-automated,zhang-etal-2022-syngec}. Arabic dependency parsing is challenging, due to its rich morphology, complex agreement patterns, and high ambiguity \citep{Marton:2013:dependency}. 
Large-scale Arabic treebanks have primarily focused on Modern Standard Arabic (MSA) and Classical Arabic (CA). The Columbia Arabic Treebank (CATiB) \citep{Habash:2009:catib} and the Penn Arabic Treebank (PATB) \citep{Maamouri:2004:penn} cover MSA, while the Camel Treebank (CamelTB) covers both MSA and CA \citep{habash-etal-2022-camel}.
Coverage of Dialectal Arabic (DA) treebanks remains uneven across varieties, with the majority focused on Egyptian Arabic \citep{maamouri-etal-2006-developing}, and 
to a lesser extent Gulf and Levantine \citep{faroukzakariaelshabrawy-etal-2026-parsing,maamouri-etal-2006-developing}.

While previous work on parallel treebanks includes the Parallel Universal Dependencies (PUD) treebank and recent parallel treebanks for closely related Turkic languages \citep{zeman-etal-2017-conll,akhundjanova-etal-2025-parallel}, no large-scale resource provides full, consistently annotated dependency trees for parallel sentences across Arabic dialects.

\begin{figure*}[th!]
\begin{center}
\includegraphics[width=1.5\columnwidth]{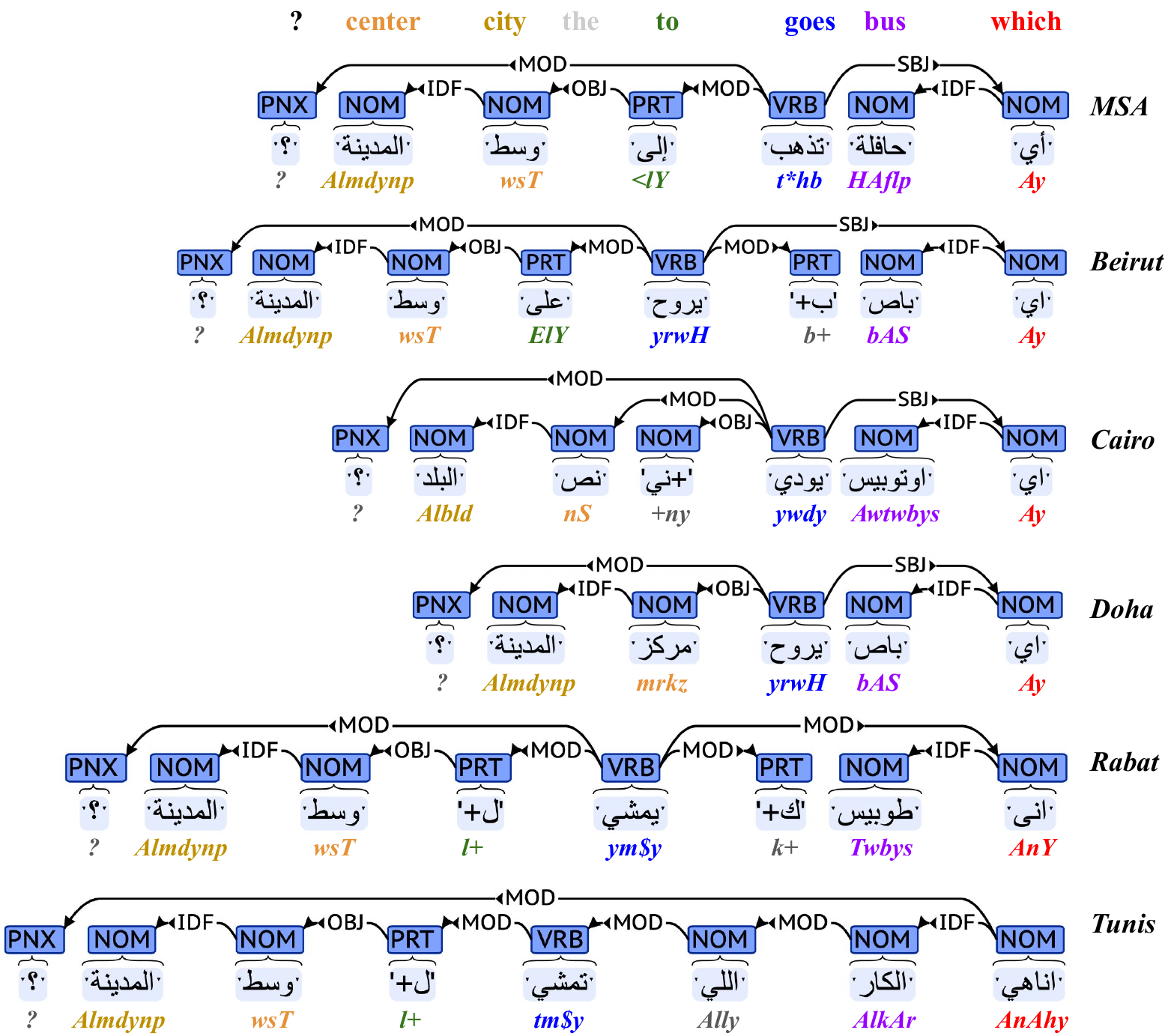}
 \caption{CATiB-style \cite{Habash:2009:syntactic} parallel dependency annotations in MSA and five city dialects. English gloss: `Which bus goes to the city center?'. Transliterations in the Buckwalter Scheme~\cite{Buckwalter:2002:buckwalter}.}
\label{fig:all_dialects_dep_trees}
\end{center}
\end{figure*}

In this paper, we introduce a first-of-its-kind linguistic resource: a parallel treebank encompassing five major Arabic city dialects, Beirut, Cairo, Doha, Rabat, and Tunis, paired with parallel MSA sentences \cite{bouamor-etal-2018-madar,Taji:2018:arabic}, all in a unified dependency representation.
This parallel architecture provides a unique opportunity to (a) quantitatively investigate the degree of syntactic similarity and divergence among MSA and its dialects while controlling for semantics; 
and (b) study cross-dialect transfer in syntactic modeling.

Our contributions can be summarized as follows:
\begin{itemize}[
  itemsep=1pt,
  parsep=0pt,
  topsep=1pt,
  partopsep=0pt
]   
    \item We introduce {\cmtb}, a newly manually annotated 5,000-sentence parallel dependency treebank spanning five Arabic city dialects, and extending a previously annotated MSA parallel version for a total of 6,000 sentences.
    \item We provide the first controlled syntactic comparison of MSA and these dialects, revealing high global similarity but local divergence.
   \item We benchmark supervised and few-shot LLM parsing, showing that pooled supervised training performs best overall and establishes new SOTA results on several parsing datasets.
\end{itemize}

\section{Linguistic Background}
\label{sec:ling}
Arabic comprises closely related linguistic varieties.  MSA is used in formal communication but has no native speakers \cite{Azzoug_2010,Ferguson:1959:diglossia}. Regional dialects (DA) dominate daily communication and increasingly appear in informal and online writing. In formal or cross-regional settings, speakers may shift toward MSA or more widely comprehensible dialect forms \cite{IriarteDiez2023whatiswhitearabic,shaban-habash-2025-arabic}. MSA and DA share a common linguistic core at the lexical, morphological, and syntactic levels, but systematic differences occur at each level. Figure~\ref{fig:all_dialects_dep_trees} illustrates this overlap and variation through parallel dependency trees.
\vspace{-1mm}
\paragraph{Lexical variation}
Some lexical items are shared across several varieties, e.g., Figure~\ref{fig:all_dialects_dep_trees}
\<المدينة> \textit{Almdynp} `city' recur in multiple varieties, while Cairo uses \<البلد> \textit{Albld}. Other words are significantly different, e.g. the word for `bus', is \<باص> \textit{bAS} (Beirut and Doha), \<الكار> \textit{AlkAr} (Tunis), \<طوبيس> \textit{Twbys} (Rabat), and\<اوتوبيس> \textit{Awtwbys} (Cairo). 
These examples illustrate the broader lexical diversity documented in MADAR, where Arabic varieties have an average pairwise lexical similarity of  25.8\% across the full dataset \citep{bouamor-etal-2018-madar}.

\vspace{-0.3mm}
\paragraph{Morphological variation}
Arabic varieties share a core morphological base, including morphemes such as feminine suffix \<ة>~\textit{p} (\textit{Ta Marbuta}), e.g., \<المدينة> \textit{Almdynp} `city' (Figure~\ref{fig:all_dialects_dep_trees}). Many verbal person-gender-number affixes are shared, but some of their realizations and distributions vary systematically, e.g., the verbal prefix -\<ن> \textit{n-} in \<نكتب> \textit{nktb} marks first person singular Tunis and Rabat `I~write', and first person plural in MSA, Beirut, Cairo, and Doha `we write'.

Arabic varieties have many pre-verbal particles that mark mood and tense. In Figure~\ref{fig:all_dialects_dep_trees}, Beirut marks the habitual present with \<ب> \textit{b+} in \<ب+ يروح> \textit{b+yrwH} `goes', whereas Rabat uses \<ك> \textit{k+} in \<ك+ يمشي> \textit{k+ym\$y}. 
The \<ب> \textit{b+} proclitic marks the future in Doha and contrasts with +\<س> \textit{s+} (MSA), +\<ه> \textit{h+} (Cairo), +\<ح> \textit{H+} (Cairo/Beirut), and +\<غ> \textit{g+} (Rabat). 

\paragraph{Syntactic variation}
The broad inventory of syntactic constructions is highly shared across Arabic varieties. Common structures include \textit{Idafa} (IDF) possessive construction, subject--verb--object configurations, and noun--adjective order. For example, despite the lexical differences in Figure~\ref{fig:all_dialects_dep_trees}, all six trees use an \textsc{IDF} relation between the two nouns expressing `center \textit{of} city': \<وسط المدينة> \textit{wsT Almdynp}, \<مركز المدينة> \textit{mrkz Almdynp}, or \<نص البلد> \textit{nS Albld}. The trees  retain a common syntactic construction even when the lexical items differ.

Syntactic variation is often construction-specific. For example, Egyptian Arabic commonly uses a postnominal demonstrative, as in \<العنوان ده> \textit{AlEnwAn dh} `this address', whereas Beirut and Doha may use the prenominal clitic \<ه> \textit{h+}, as in \<هـ+ العنوان> \textit{h+AlEnwAn}. Preposition choice and argument structure also vary. In Figure~\ref{fig:all_dialects_dep_trees}, destinations take \<إلى> \textit{<lY} `to' in MSA, \<على> \textit{ElY} `at' in Beirut, and +\<ل>  \textit{l+} `to/for' in Rabat and Tunis. Doha uses no preposition, treating `city center' as a direct nominal argument. Cairo instead uses \<يودي> \textit{ywdy} `take' with an object clitic, yielding a different argument structure.

\section{Related Work}
\label{sec:related}

\paragraph{Arabic Treebanks}
Arabic syntactic annotation has historically focused on MSA and CA.
PATB combines rich morphological annotation with phrase-structure trees \cite{Maamouri:2004:penn,Maamouri:2008:enhancing}.
The Prague Arabic Dependency Treebank (PADT) introduced a multi-layer dependency representation for Arabic \cite{Hajic:2004:prague}, while CATiB was annotated directly using dependency formalism inspired by traditional Arabic grammar \cite{Habash:2009:catib}.
Additional resources include the Quranic Arabic Dependency Treebank \cite{dukes-etal-2010-syntactic}, the Arabic Basic Traveling Expressions Corpus \cite{Taji:2018:arabic}, and CamelTB which consists of MSA and CA across 13 subcorpora \cite{habash-etal-2022-camel}.

Beyond MSA and CA, several treebanks have been developed for individual Arabic dialects.
Early Linguistic Data Consortium (LDC) efforts produced treebanks for Levantine~\cite{maamouri-etal-2006-developing} and Egyptian Arabic \cite{Maamouri:2012:arz-4,maamouri-etal-2014-developing}.
More recently, NArabizi provided UD annotation for romanized, Algerian Arabic--French texts \cite{seddah-etal-2020-building}, while \citet{faroukzakariaelshabrawy-etal-2026-parsing} released a CATiB dependency treebank for Gulf Arabic.

Most existing dialectal treebanks have been developed for individual dialects, with fewer efforts covering multiple dialects within a single annotation framework.
The closest work to ours is \citet{mokh-etal-2024-domain}, who annotated 200 parallel sentences in four Arabic dialect groups using CATiB-style dependencies with a focus on two specific linguistic constructions. Their data is not publicly available. Our work differs in both scale and coverage, providing uniform CATiB dependency annotations for 5,000 parallel sentences across five dialects.

\paragraph{Arabic dependency parsing.}

Arabic dependency parsing has primarily focused on MSA.
Methods range from transition-based parsing \cite{Habash:2009:catib,Marton:2013:dependency,Shahrour:2016:camelparser} to multi-task learning \cite{kankanampati-etal-2020-multitask}, graph-based dependency parsing \cite{elshabrawy-etal-2023-camelparser2}, and more recently, large language models \cite{adel2026arabic}.
Dialectal Arabic parsing has received comparatively less attention.
Early work focused on adapting MSA resources through treebank transduction \cite{chiang-etal-2006-parsing} and multilingual transfer~\cite{muller2020multilinguallanguagemodelstransfer}.
More recently, \citet{mokh-etal-2024-domain} evaluated an MSA-trained parser on a small parallel corpus spanning four Arabic dialects, while \citet{faroukzakariaelshabrawy-etal-2026-parsing} developed a multi-variety Arabic parser that covers Egyptian and Gulf.
In this work, we benchmark supervised and few-shot LLM parsing and establish new SOTA results on several parsing datasets.

\section{The {\cmtb} Treebank}
\label{sec:treebank}

\subsection{Source Data}
We selected 1,000 aligned sentences for Beirut, Cairo, Doha, Rabat, and Tunis from MADAR-CODA \citep{eryani-etal-2020-spelling}, which draws its sentences from MADAR \citep{bouamor-etal-2018-madar} and uses the Conventional Orthography for Dialectal Arabic (CODA) \cite{habash-etal-2018-unified}. We obtained the corresponding MSA sentences from CamelTB \citep{habash-etal-2022-camel,Taji:2018:arabic}.   This parallel structure allows us to compare syntactic similarities and differences across the five dialects and MSA.

\subsection{Annotation Scheme}
Following CamelTB's annotation of the MADAR MSA data, we adopt the CATiB annotation scheme \citep{Habash:2009:syntactic,habash-etal-2022-camel}, an Arabic-specific dependency representation inspired by traditional Arabic syntax.
Unlike Universal Dependencies (UD), which uses a shared multilingual scheme to prioritize cross-linguistic comparability \citep{de2021universal}, CATiB captures language-specific structures.
As a functional-head dependency representation, CATiB is similar to Surface-Syntactic Universal Dependencies (SUD) \citep{Gerdes:2021:SUD}. We summarize the scheme next \cite{Habash:2009:syntactic}.

\paragraph{Tokenization}
CATiB follows the Penn Arabic Treebank tokenization guidelines \cite{Maamouri:2004:penn}; segment all clitics except the definite article \<ال> \textit{Al+} `the', and normalize basewords to their uncliticized form.  For example, the word \<وبمدينتنا> \textit{wbmdyntnA} 
tokenizes as \<نا>+
\<مدينة>
+\<ب>
+\<و>
\textit{w+}~'and'~\textit{b+}~'in'~\textit{mdynp}~`city' \textit{+nA}~`our'.

\paragraph{Parts-of-Speech}
CATiB uses six {\pos} tags: {\textit{NOM}} (nominals: nouns, adjectives), {\textit{PROP}} (proper nouns), {\textit{VRB}} (active verbs), {\textit{VRB-PASS}} (passive verbs), {\textit{PRT}} (particles, including prepositions and conjunctions), and {\textit{PNX}} (punctuation). This tagset facilitates a balance of economic annotation while preserving the level of syntactic detail required for this study. 

\paragraph{Dependency relations} CATiB uses eight relation labels: \textit{SBJ} (subjects of verbs in SVO and VSO constructions, and topics of simple nominal sentences), \textit{OBJ} (objects of verbs, prepositions, conjunctions, or deverbal nouns), \textit{TPC} (topics of complex nominal sentences with explicit pronominal reference), \textit{PRD} (predicative complements in extended copular constructions), \textit{IDF} (the \textit{idafa} possessive construction), \textit{TMZ} (the \textit{tamyiz} specification construction), \textit{MOD} (general modification of verbs or nominals), and \textit{---} (flat constructions, such as multiword proper names, and multiple roots in sentences containing more than one full clause).
\subsection{Annotation Process}

\begin{table*}[t]
\centering
\small
\tabcolsep3.5pt
\begin{tabular}{lrrrrrrrrrrrr}
\toprule
& \multicolumn{3}{c}{\textbf{Train}} & \multicolumn{3}{c}{\textbf{Development}} & \multicolumn{3}{c}{\textbf{Test}} & \multicolumn{3}{c}{\textbf{Total}} \\
\cmidrule(lr){2-4} \cmidrule(lr){5-7} \cmidrule(lr){8-10} \cmidrule(lr){11-13}
\textbf{Variety}  & \textbf{Sent.} & \textbf{Words} & \textbf{Tokens} & \textbf{Sent.} & \textbf{Words} & \textbf{Tokens} & \textbf{Sent.} & \textbf{Words} & \textbf{Tokens} & \textbf{Sent.} &  \textbf{Words} & \textbf{Tokens} \\
\midrule
\textbf{Beirut} & 700 & 3,642 & 5,843 & 150 & 803 & 1,312 & 150 & 840 & 1,307 & 1,000 & 5,285 & 8,462 \\
\textbf{Cairo} & 700 & 3,987 & 5,973 & 150 & 898 & 1,342 & 150 & 899 & 1,328 &1,000 & 5,784 & 8,643 \\
\textbf{Doha} & 700 & 3,625 & 5,369 & 150 & 802 & 1,204 & 150 & 809 & 1,195 &1,000 & 5,236 & 7,768 \\
\textbf{Rabat} & 700 & 4,125 & 6,185 & 150 & 900 & 1,367 & 150 & 925 & 1,373 &1,000 & 5,950 & 8,925 \\
\textbf{Tunis} & 700 & 3,851 & 5,845 & 150 & 844 & 1,322 & 150 & 864 & 1,302 &1,000 & 5,559 & 8,469 \\
\textbf{MSA} & 700 & 5,487 & 6,395 & 150 & 1,263 & 1,480 & 150 & 1,208 & 1,419 & 1,000 &7,958 & 9,294 \\
\midrule
\textbf{All Dialects} & 3,500 & 19,230 & 29,215 & 750 & 4,247 & 6,547 & 750 & 4,337 & 6,505 &5,000 & 27,814 & 42,267 \\
\textbf{All Dialects + MSA} & 4,200 & 24,717 & 35,610 & 900 & 5,510 & 8,027 & 900 & 5,545 & 7,924 &6,000 & 35,772 & 51,561 \\
\bottomrule
\end{tabular}
\caption{Numbers of sentences, words, and tokens in the train,
development, and test splits. The six varieties are parallel at the
sentence level.}
\label{tab:treebank-splits}
\end{table*}

\paragraph{Automatic Annotation}

First, we automatically tokenize the data and  assign POS tags using the Egyptian morphological disambiguation model in Camel Tools~\cite{Obeid:2020:camel}.
We then obtain initial parses using a biaffine parser (similar to \citet{elshabrawy-etal-2023-camelparser2}) trained on CamelTB~\cite{habash-etal-2022-camel}, PATB~\cite{Maamouri:2004:penn,Maamouri:2008:enhancing}, and ARZTB~\cite{maamouri-etal-2014-developing}, converted from constituency to dependency using Arabic-C2D \cite{camellab_arabic_c2d}.

\paragraph{Human Validation}
Annotating multiple Arabic dialects requires expertise in dialectal variation and syntactic analysis.
Our team comprised (i) a primary annotator, a Sudanese Arabic speaker with over 18 years of treebank annotation experience; (ii) a secondary annotator, a Palestinian Arabic speaker and senior computational linguist with 25 years of experience; (iii) and a validator, a Palestinian Arabic speaker and trained computational linguist specializing in comparative dialectology.
Native speakers of the five varieties were consulted as needed. The parallel translations in English, MSA, and the other dialects were also used as references during annotation.

The annotation team made minor extensions to the annotation guidelines to accommodate dialect-specific lexical variants while preserving CATiB's general principles, e.g., extending the tokenizable clitic set to include negation and verbal particles. The primary annotator first corrected the automatically generated {\pos} tags and dependency annotations. The validator, assisted by the secondary annotator, then reviewed and corrected the full corpus. We report the resulting quality statistics below.

\paragraph{Quality Statistics}

The \textit{initial annotation round} required extensive corrections to the automatic output. 
The amount of edits in this round can be quantified by comparing the corrected annotations to the automatic output:  93.6\% tokenization F$_1$, 84.8\%  {\pos} accuracy, and 77.6\% labeled attachment score (LAS). The \textit{validation round} only identified minor changes to the initial annotations. One example is the predicating active participle \<عايز> \textit{EAyz} `wanting/I want' (Cairo) which was corrected from `VRB' to `NOM'. The validation results can be similarly quantified: 99.1\% tokenization F$_1$, 99.6\% {\pos} accuracy, and 99.5\% LAS.  These results indicate strong consistency between the two rounds and support the quality of the final annotations.

\subsection{Corpus Statistics}
\label{sec:corpus-statistics}

{\cmtb} comprises 1,000 parallel sentence sets: 5,000 newly annotated sentences from Beirut, Cairo, Doha, Rabat, and Tunis, aligned with their 1,000 MSA counterparts from CamelTB \citep{habash-etal-2022-camel,Taji:2018:arabic}. We split the parallel sets into 70\% training, 15\% development, and 15\% test data, keeping all aligned realizations in the same split. Table~\ref{tab:treebank-splits} reports the sentence, word, and token counts. Additional statistics are provided in Appendix Tables~\ref{tab:treebank-descriptive-statistics}--\ref{tab:treebank-relations}.

\section{Dialect Syntax Similarity}
The sentence-parallel design of {\cmtb} enables controlled syntactic comparison across varieties. We compare global construction distributions and local analyses over aligned tokens, then examine a number of specific canonical constructions. 

\subsection{Global Construction Profiles} 
We represent every variety's 
trees by counts of the ordered construction $\langle$head POS, dependency relation, dependent POS, direction$\rangle$, where direction distinguishes dependents before and after their heads and root arcs receive a separate value.  The union inventory contains 150 constructions. Pairwise cosine similarities (Table~\ref{tab:cosine-similarity}) are uniformly high
(93.3--98.3\%), indicating a shared constructional core. The closest pair is Rabat--Tunis (Maghreb region), followed by Doha--Tunis (Gulf--Maghreb) and Cairo--Beirut (Egypt--Levant) suggesting that similarity  is not determined by geography alone.

\begin{table*}[t]
\centering
\begin{minipage}[t]{0.485\textwidth}
\vspace{0pt}
\centering
\captionsetup{hypcap=false}
\small
\setlength{\tabcolsep}{5pt}
\begin{tabular}{@{}lrrrrrr@{}}
\toprule
 & \multicolumn{6}{c}{\textbf{Source}}\\ \cmidrule{2-7}
\textbf{Target} & \textbf{Beirut} & \textbf{Cairo} & \textbf{Doha} & \textbf{Rabat} & \textbf{Tunis} & \textbf{MSA} \\
\midrule
Beirut & -- & \textbf{97.5} & 96.3 & 96.3 & 94.7 & 96.8 \\
Cairo  & \textbf{97.5} & -- & 96.8 & 94.5 & 93.8 & 93.5 \\
Doha   & 96.3 & 96.8 & -- & 97.2 & \textbf{97.5} & 93.3 \\
Rabat  & 96.3 & 94.5 & 97.2 & -- & \textbf{98.3} & 94.9 \\
Tunis  & 94.7 & 93.8 & 97.5 & \textbf{98.3} & -- & 94.9 \\
MSA    & \textbf{96.8} & 93.5 & 93.3 & 94.9 & 94.9 & -- \\
\bottomrule
\end{tabular}

\captionof{table}{Global construction-count cosine similarity (\%).
Bold marks each row's highest off-diagonal value.}
\label{tab:cosine-similarity}
\end{minipage}\hfill
\begin{minipage}[t]{0.485\textwidth}
\vspace{0pt}
\centering
\captionsetup{hypcap=false}
\small
\setlength{\tabcolsep}{5pt}
\begin{tabular}{@{}lrrrrrr@{}}
\toprule
  & \multicolumn{6}{c}{\textbf{Source}}\\ \cmidrule{2-7}
\textbf{Target} & \textbf{Beirut} & \textbf{Cairo} & \textbf{Doha} & \textbf{Rabat} & \textbf{Tunis} & \textbf{MSA} \\
\midrule
Beirut & -- & \textbf{81.6} & 76.0 & 70.8 & 70.1 & 70.7 \\
Cairo  & \textbf{81.6} & -- & 78.1 & 66.7 & 66.8 & 70.1 \\
Doha   & 76.0 & \textbf{78.1} & -- & 75.6 & 73.7 & 77.9 \\
Rabat  & 70.8 & 66.7 & 75.6 & -- & \textbf{83.8} & 74.8 \\
Tunis  & 70.1 & 66.8 & 73.7 & \textbf{83.8} & -- & 74.6 \\
MSA    & 70.7 & 70.1 & \textbf{77.9} & 74.8 & 74.6 & -- \\
\bottomrule
\end{tabular}
\captionof{table}{Complete agreement (Token POS, Head POS, Relation, Direction) among aligned token pairs (\%).}
\label{tab:complete-agreement}
\end{minipage}
\end{table*}

\subsection{Alignment-Level Similarity}

\paragraph{Alignment Pipeline and Manual Evaluation}
Because alignment-level similarity analysis requires one-to-one token alignments, we generate them automatically using Awesome-Align~\cite{dou-neubig-2021-word}. We initialize the model with CAMeLBERT-Mix~\cite{inoue-etal-2021-interplay}, which is pretrained on both MSA and DA, and fine-tune it on all parallel sentence pairs in the treebanks. We manually evaluate alignment quality on a random sample of 20 sentence pairs from each of the 15 unordered cross-variety pairs, for a total of 300 sentence pairs. Two of our lead annotators jointly inspected all token alignments in the sample. The automatically generated alignments achieve 93.4\% recall and  92.2\% F$_1$ against the manual sample.\footnote{Appendix Table~\ref{tab:alignment-hparams} reports the full alignment and training details, along with comparisons to two alternative alignment systems that we considered: SimAlign~\cite{jalili-sabet-etal-2020-simalign} and MGIZA++~\cite{Gao:2008:parallel}.}

For every sentence and variety pair, we convert the Awesome-Align links
into a maximum-cardinality one-to-one matching, breaking ties by summed
alignment probability.  This retains 107,158 of 111,251 original links
(96.3\%).  Across the 15 unordered variety pairs, an average of 83.3\%
of tokens are aligned. Of the selected token pairs, 69,104 (64.5\%)
also have aligned heads. These head-aligned token pairs form the
evaluation set for the agreement measures below.

\paragraph{Alignment-level Complete Agreement}

Our primary measure is complete agreement.  For an aligned token
pair whose heads also align, it requires four decisions to agree
simultaneously: the token POS, head POS, dependency relation, and
dependency direction.  It is therefore a strict test of whether the two
tokens receive the same local syntactic configuration.  Table~\ref{tab:complete-agreement} reports this measure for every variety pair.

Among head-aligned token pairs, complete agreement averages 74.1\%
across the 15 variety pairs and ranges from 66.7\% to 83.8\%.
Rabat--Tunis has the highest agreement, followed by Beirut--Cairo;
Cairo--Rabat has the lowest.  Doha--MSA is the strongest MSA pairing.
Within the same head-aligned set, token-and-head POS matches in 78.7\%
of cases, compared with 91.6\% for dependency relation and 98.7\% for
direction.  POS configuration therefore accounts for most of the
remaining disagreement (Appendix
Table~\ref{tab:appendix-alignment}).

\subsection{Syntactic Structure Analysis} 

We further investigate the most frequent patterns across dialects, to understand how relations are realized. 
Below, we provide our observations on three constructions: (1) sentential constructions, (2) idafa construction, and (3) Root head. 

\paragraph{Sentential Constructions}
We distinguish three constructions: (a) Verb-Subject Construction (VS) (b) Subject-Verb Construction (SV) (c) Verbless constructions (noV). MSA strongly favors VS whereas all five dialects favor SV order. noV constructions are more common in dialects (63--91 times per dialect vs. 26 in MSA), as in the example \<عندي حساسية ديال القطوط> \textit{Endy HsAsyp dyAl AlqTwT} `I have cats allergy' in Rabat, where the predicate is headed by a noun.

\paragraph{Idafa Constructions} We focus on analytical Idafa. We observe that parallel Idafa constructions are realized differently across dialects. In Doha, Rabat and Tunis, analytic Idafa is more frequently realized using the relational nouns
\<حق> \textit{Haq}
,\<ديال> \textit{dyAl},
and \<متاع> \textit{mtA{\AYN}}
`of/for', respectively. In the parallel constructions, in Cairo and Beirut, Idafa is realized sometimes with a relational noun such as
\<بتاع> \textit{btA{\AYN}} and \<تبع> \textit{mtA{\AYN}}, and other times 
with the preposition +\<ل> \textit{l}+ `for/to', resulting in  different dependency structures.

\paragraph{Root Head} Finally, we compare the POS categories of tokens attached directly to ROOT. Verbs are the most frequent ROOT dependents in MSA, Rabat, and Tunis, while nominals are most frequent in Cairo and Beirut; Doha is nearly balanced. The higher frequency of nominal ROOT dependents in Cairo and Beirut may partly reflect nominally annotated `desire' expressions, e.g. \<عايز> \textit{EAyz} (Cairo) and \<بدي> \textit{bdy} (Beirut), which appear as verbal in other varieties: \<أريد> \textit{{>ryd}} (MSA) or \<أبغي> \textit{>bgy} (Doha),  and \<نحب> \textit{nHb} (Tunis and Rabat).
 
\section{Parsing Benchmark}
\label{sec:approach}

\paragraph{Data}
We use {\cmtb} partitions for the city dialects and MSA (Table~\ref{tab:treebank-splits}).
Sentence IDs are split jointly across varieties, so no translation of an evaluation sentence appears in training. We also use four external CATiB-style treebanks: PATB~\citep{Maamouri:2004:penn} and ARZTB~\citep{maamouri-etal-2014-developing}, converted with Arabic-C2D~\citep{camellab_arabic_c2d} and split following \citet{diab2013ldcarabictreebanksassociated}; and CamelTB~\citep{habash-etal-2022-camel}\footnote{Since we use the MADAR MSA sentences in CamelTB, we excluded our 300 dev and test sentences to avoid leakage.} and CamelTB-Gumar~\citep{faroukzakariaelshabrawy-etal-2026-parsing}, using their published splits. We report each variety separately, the five dialects together, and all six varieties together, on development and test.

\paragraph{Metric}
We report Labeled Attachment Score (LAS), the percentage of tokens assigned both the correct head and dependency relation.
We use development LAS for model selection and reserve the test set for the final comparison. We assess system differences with paired bootstrap 95\% confidence intervals, resampling by sentence ID to preserve the parallel structure (Appendix Table~\ref{tab:bootstrap-parsing}).

\paragraph{Supervised Parsers}

All supervised models (henceforth, \textbf{Supervised}) follow the CAMeLParser-Dialects setup \citep{faroukzakariaelshabrawy-etal-2026-parsing}: a CAMeLBERT-Mix encoder \citep{inoue-etal-2021-interplay} feeds the biaffine parser of \citet{dozat2017deepbiaffineattentionneural}, implemented in SuPar \citep{zhang-etal-2020-efficient}. 
However, unlike CAMeLParser-Dialects, we use projective multi-root decoding instead of  single-root decoding following \citep{adel2026arabic}.\footnote{
This minor modification matters because 191 of 900 development sentences (21.2\%) contain more than one root, and \citet{adel2026arabic}
identified the same single-root limitation in CamelParser decoding on CamelTB. Detailed comparison results are in Appendix Table~\ref{tab:external-treebank-test}.}

We train eight {\cmtb} models: one on each variety, one on the five pooled dialects, and one on all six varieties.
We also train models on PATB+CamelTB, on all four external treebanks, and on the four external treebanks plus the full {\cmtb} training set.  The first two external combinations match the data used by CamelParser2.0 and CAMeLParser-Dialects, respectively, but differ in decoding (multi-root not single-root).

\paragraph{LLM Parsers}
Our evaluation includes closed-source multilingual models, open-weight multilingual models, and Arabic-centric models: Gemini-3.6-Flash~\cite{gemini}, GPT-5.6-Sol~\cite{openai2026gpt56}, Llama-4-Scout~\cite{meta2025llama4}, and Fanar-2~\cite{fanar2}, henceforth \textbf{Gemini}, \textbf{GPT}, \textbf{Llama}, and \textbf{Fanar}, respectively. Following \citet{adel2026arabic}, the prompt defines the CATiB relations and asks for one head--relation prediction per token.
Gemini and GPT use their structured-output interfaces; Llama and Fanar use vLLM structured decoding~\citep{kwon2023efficient}. We test zero-shot prompting and retrieval. Following \citet{adel2026arabic}, for each evaluation sentence, we add to the prompt the ten {\cmtb} training trees whose sentences are most lexically similar to the specific evaluation sentence using chrF++ scores~\citep{popovic-2017-chrf}.

\section{Results}
\label{sec:eval}
Table~\ref{tab:parsing-dev} reports the complete development matrix used for model selection.
Table~\ref{tab:parsing-test} compares the selected systems on the test set, and Appendix Table~\ref{tab:appendix-parsing-test} gives the full test matrix.

\begin{table*}[t]
\centering
\small
\setlength{\tabcolsep}{4pt}
\begin{tabular}{llcccccccc}
\toprule
\textbf{Model}
& \textbf{Training Data / Prompting}\phantom{XXXXXXXX}
& \textbf{Beirut}
& \textbf{Cairo}
& \textbf{Doha}
& \textbf{Rabat}
& \textbf{Tunis}
& \textbf{MSA}
& \textbf{Dialects}
& \textbf{All} \\
\midrule

\textbf{Supervised}
& {\cmtb} Beirut
& 81.1 & 77.8 & 78.7 & 73.0 & 75.0 & 72.7 & 77.1 & 76.3 \\

\textbf{Supervised}
& {\cmtb} Cairo
& 73.2 & 82.0 & 82.2 & 68.0 & 74.8 & 75.6 & 75.9 & 75.8 \\

\textbf{Supervised}
& {\cmtb} Doha
& 73.4 & 78.9 & 85.4 & 70.2 & 76.5 & 73.7 & 76.7 & 76.1 \\

\textbf{Supervised}
& {\cmtb} Rabat
& 66.5 & 72.9 & 76.2 & 83.9 & 75.1 & 70.1 & 75.0 & 74.1 \\

\textbf{Supervised}
& {\cmtb} Tunis
& 66.0 & 73.7 & 77.4 & 71.0 & 83.4 & 71.9 & 74.2 & 73.8 \\

\textbf{Supervised}
& {\cmtb} MSA
& 64.6 & 67.2 & 70.8 & 66.6 & 67.9 & 83.2 & 67.4 & 70.3 \\

\midrule
\textbf{Supervised}
& {\cmtb} Dialects
& 85.4 & 85.9 & 88.4 & 87.7 & 87.6 & 80.6 & 87.0 & 85.8 \\

\textbf{Supervised}
& {\cmtb} All
& 84.5 & 86.1 & 88.9 & 86.5 & 86.6 & 85.5 & 86.5 & 86.3 \\

\midrule
\textbf{Supervised}
& PATB + CamelTB
& 74.2 & 77.9 & 81.0 & 73.3 & 76.9 & 91.0 & 76.5 & 79.2 \\

\textbf{Supervised}
& PATB + CamelTB + ARZTB + Gumar
& 77.9
& \textbf{88.3}
& 89.5
& 74.3
& 81.1
& \textbf{91.1}
& 82.0
& 83.7 \\

\midrule
\textbf{Supervised}
& All Treebanks
& \textbf{86.7}
& 88.0
& \textbf{92.5}
& \textbf{89.0}
& \textbf{89.3}
& 88.2
& \textbf{89.0}
& \textbf{88.9} \\

\midrule\midrule
\textbf{Gemini}
& 0-shot
& 68.4 & 72.3 & 70.6 & 72.3 & 72.5 & 75.3 & 71.3 & 72.0 \\

\textbf{Gemini}
& 10-shot
& \underline{86.2}
& \underline{85.6}
& \underline{89.2}
& \underline{88.9}
& \underline{88.4}
& \underline{86.1}
& \underline{87.6}
& \underline{87.3} \\

\midrule
\textbf{GPT}
& 0-shot
& 64.1 & 67.5 & 66.4 & 67.1 & 65.4 & 70.9 & 66.1 & 67.0 \\

\textbf{GPT} 
& 10-shot
& 84.2 & 83.9 & 86.0 & 86.1 & 86.1 & 84.3 & 85.2 & 85.1 \\

\midrule
\textbf{Fanar}
& 0-shot
& 14.8 & 13.7 & 14.5 & 16.2 & 16.0 & 18.9 & 15.0 & 15.8 \\

\textbf{Fanar}
& 10-shot
& 73.4 & 72.5 & 76.3 & 74.5 & 74.4 & 76.2 & 74.2 & 74.6 \\

\midrule
\textbf{Llama}
& 0-shot
& 22.7 & 24.5 & 24.7 & 22.6 & 23.0 & 29.7 & 23.5 & 24.6 \\

\textbf{Llama}
& 10-shot
& 70.0 & 70.3 & 74.0 & 70.9 & 72.6 & 75.5 & 71.5 & 72.3 \\

\bottomrule
\end{tabular}
\caption{
Development-set parsing results (LAS) for supervised parsers and prompted
LLMs.
All \textbf{Supervised} rows use the CAMeLBERT-Mix Biaffine  model and projective
multi-root decoding; their configurations differ only in training data.
All Treebanks denotes PATB, CamelTB, ARZTB, Gumar, and {\cmtb}.
Bold indicates the best result overall, and underlining indicates the best LLM
result.
}
\label{tab:parsing-dev}
\end{table*}

\begin{table*}[t]
\centering
\small
\setlength{\tabcolsep}{4pt}
\begin{tabular}{llcccccccc}
\toprule
\textbf{Model}
& \textbf{Training Data / Prompting}\phantom{XXXXXXXX}
& \textbf{Beirut}
& \textbf{Cairo}
& \textbf{Doha}
& \textbf{Rabat}
& \textbf{Tunis}
& \textbf{MSA}
& \textbf{Dialects}
& \textbf{All} \\
\midrule

\textbf{Supervised} & All treebanks
& \textbf{86.9}
& \textbf{87.4}
& \textbf{90.7}
& \textbf{85.4}
& 87.9
& \textbf{86.9}
& \textbf{87.6}
& \textbf{87.5} \\

\midrule
\textbf{Gemini} 
& 10-shot
& \underline{85.8}
& \underline{83.3}
& \underline{86.9}
& \underline{83.8}
& \textbf{\underline{88.1}}
& \underline{86.4}
& \underline{85.5}
& \underline{85.7} \\

\textbf{GPT}
& 10-shot
& 84.5 & 81.9 & 84.5 & 82.4 & 85.0 & 85.1 & 83.6 & 83.9 \\

\textbf{Fanar}
& 10-shot
& 67.6 & 70.0 & 72.3 & 72.6 & 72.5 & 73.3 & 71.0 & 71.4 \\

\textbf{Llama}
& 10-shot
& 63.4 & 66.9 & 70.2 & 67.5 & 69.0 & 69.4 & 67.3 & 67.7 \\

\bottomrule
\end{tabular}
\caption{
Test-set parsing results (LAS) for the best supervised configuration and the best setup for each LLM, selected on development data. ``All treebanks'' denotes PATB, CamelTB, ARZTB, Gumar, and the proposed {\cmtb}. Bold indicates the best result overall, and underlining indicates the best LLM result. The complete test matrix appears in Appendix Table~\ref{tab:appendix-parsing-test}.
}
\label{tab:parsing-test}
\end{table*}

\subsection{Supervised Parsing}

\paragraph{Pooling Dialects is Essential}
On development data, single-variety parsers are almost always best on their own variety, even though the sentences are parallel.
However, they transfer poorly: the best single-variety model reaches only 77.1 LAS across the five dialects.
Pooling the five dialects raises this score to 87.0, a gain of 9.9 points (95\% CI: 8.6--11.2).
Thus, dialect differences still matter for parsing even when the sentences express the same content; pooling significantly outperforms every single-dialect model.

\paragraph{MSA and Dialect Data are Complementary}
On development data, the MSA-only model reaches 67.4 LAS on the dialects, while the single-dialect models reach only 70.1--75.6 on MSA.
Adding MSA sentences to the pooled dialect training data raises MSA LAS from 80.6 to 85.5, while dialect LAS drops by only 0.5 points.
Across all six varieties, the score rises by 0.5 points, but this gain is not significant. A joint model can therefore cover MSA and the five dialects without giving up much dialect performance.

\paragraph{Direct Dialect Coverage Gives the Strongest Supervised Model}
On development data, PATB and CamelTB alone reach 76.5 LAS on the dialects.
Adding the Egyptian ARZTB and Gulf Gumar treebanks raises this to 82.0.
Adding the proposed {\cmtb} raises it again to 89.0 and gives the best supervised result for four of the five dialects.
Cairo is the exception, with an essentially unchanged score (88.3 to 88.0).
The gains are much larger for Beirut, Rabat, and Tunis (8.8, 14.7, and 8.2 points), which are not covered by the external treebanks.
The all-treebanks model is 5.2 points better overall than the four-external-treebank model and
2.6 points better than training on the proposed treebank alone.
Its only cost is a 2.9-point drop on MSA relative to the four-external-treebank model. 

\subsection{LLM Parsing}

\paragraph{Retrieval Makes LLMs Competitive}

On development data, ten examples improve Gemini from 72.0 to 87.3 LAS and GPT from 67.0 to 85.1. The gains are larger for Fanar (15.8 to 74.6) and Llama (24.6 to 72.3), and the improvements are significant for all four LLMs. Gemini is the strongest LLM, but the all-treebanks supervised model remains better overall on both development (88.9 versus 87.3 LAS) and test data (87.5 versus 85.7). On the test set, Gemini is best only on Tunis (88.1 versus 87.9); the supervised model is best on the other five varieties. The largest gaps are on Cairo and Doha, where the supervised training data includes Egyptian and Gulf treebanks. Thus, retrieval makes Gemini competitive, but broad supervised coverage remains the strongest and most consistent setup.

\subsection{Performance Analysis}

We analyze performance patterns on the development set for the best supervised and LLM parsers: the all-treebanks supervised model (henceforth, \textit{Supervised}) and \textit{Gemini}. \textbf{First}, {Supervised} outperforms {Gemini} on 52 of 135 construction types, covering 73\% of gold instances, whereas {Gemini} wins on 26 types covering 25.1\%. \textbf{Second},  {Supervised} performs best on frequent core constructions, including Prep-Noun, Verb-Object, Idafa, and Verb-PP (94.2--98.7 F$_1$), and worst on nominal modification and punctuation attached to particles (77.8--84.7 F$_1$). Its F$_1$ declines from 91\% on the 16 most frequent types (80\% of instances) to 66\% on the 93 rarest types (4\%). \textbf{Third}, parent-before-child dependencies are the most common (74\%) and easiest, with {Supervised} leading in both directions by 1.5--1.8 points. \textbf{Fourth},
performance generally decreases with dependency distance, except beyond five tokens, where punctuation--verb attachments dominate. \textbf{Finally},
across dependency relations, {Supervised} outperforms {Gemini}. \textsc{OBJ} is the easiest (95.0 F$_1$; 20.8\% of instances), while \textsc{MOD} is the largest class (86.3 F$_1$; 45.5\%). The lowest-scoring relations are \textsc{TMZ} and \textsc{TPC} (62.5 and 63.6 F$_1$), but they contain only 18 and 11 instances, respectively.

\section{Conclusions and Future Work}
We introduced {\cmtb}, the first parallel multi-dialect Arabic dependency treebank.
We provided quantitative analysis of dialectal syntactic similarity across the five cities, and established the first benchmark for dialect parsing across five cities using both a supervised parser and four LLMs. 
Supervised parsers remain the strongest for DA.

Future work will explore extending {\cmtb} to additional MADAR varieties through parallel annotation projection and controlled transfer. We plan to develop multidialect tokenization for end-to-end parsing, and automatically map the annotations to UD/SUD for broader cross-linguistic comparison.

\clearpage

\section{Limitations}

{\cmtb} covers only five city varieties and only 1,000 translated travel-domain sentences per variety. Though necessary to isolate dialectal differences, its controlled parallel design may not capture the syntactic diversity of spontaneous speech, online writing, code-switching, other domains, or variation within and beyond these cities.

The local similarity analysis relies on automatic one-to-one alignments and is restricted to token pairs with aligned heads, potentially underrepresenting divergent lexicalizations, clitic structures, and implicit arguments. Moreover, CATiB limits direct comparison with other annotation frameworks (such as UD and SUD).

Finally, the parsing benchmark assumes gold tokenization and evaluates a single supervised architecture and four LLMs under fixed prompting settings; the results may not generalize to end-to-end parsing or other models and retrieval strategies.

\section*{Ethics Statement}
Our annotators were compensated fair wages at rates commensurate with the time and effort put into this annotation task. Our data is translations of sentences that contain no private or personally identifying information about the annotators or other real individuals. Our data will be released under a license that is in accordance with the licensing terms of the source dataset. We used AI writing assistance within the scope of ``Assistance
purely with the language of the paper'' described in the ACL Policy on Publication Ethics.

\bibliography{anthology,camel-bib-v3,custom}

\appendix
\section{Licenses}
\label{app:license}

We list below the licenses and terms governing the principal data, models, and
tools used in this work. All resources are used in accordance with their
intended research use and distribution conditions.

\begin{itemize}
    \item \textbf{MADAR and MADAR-CODA.}
    MADAR \cite{bouamor-etal-2018-madar} and MADAR-CODA \cite{eryani-etal-2020-spelling} are available for non-commercial research and evaluation under licenses that prohibit redistribution and modification of the original resources. We use both resources solely under these terms. 

    We will release our independently created annotation layer and code that enables licensed users to combine it with their own copies of MADAR and MADAR-CODA, subject to the original license terms.

    \item \textbf{External treebanks.}
    PATB~\cite{Maamouri:2004:penn} and ARZTB~\cite{maamouri-etal-2014-developing} are LDC
    resources and are used under the applicable LDC user agreements. CamelTB
    and CamelTB-Gumar~\cite{habash-etal-2022-camel,
    khalifa-etal-2016-large,Khalifa:2018:morphologically} are used under the
    terms accompanying their releases. We do not redistribute any of these
    third-party treebanks.

    \item \textbf{Supervised parser.}
    The CAMeLBERT-Mix checkpoint~\cite{inoue-etal-2021-interplay} is released
    under the Apache License 2.0.\footnote{\url{https://huggingface.co/CAMeL-Lab/bert-base-arabic-camelbert-mix}}
    The CamelParser2.0 parsing model~\cite{elshabrawy-etal-2023-camelparser2}
    and SuPar~\cite{zhang-etal-2020-efficient} are released under the MIT
    License.\footnote{\url{https://huggingface.co/CAMeL-Lab/camelbert-catib-parser}}
    \footnote{\url{https://github.com/yzhangcs/parser}}

    \item \textbf{Open-weight LLMs and inference.}
    Fanar-2-27B-Instruct~\cite{fanar2} is released under the Apache License
    2.0,\footnote{\url{https://huggingface.co/QCRI/Fanar-2-27B-Instruct}}
    whereas Llama-4-Scout-17B-16E-Instruct~\cite{meta2025llama4} is governed
    by the Llama 4 Community License Agreement.\footnote{\url{https://huggingface.co/meta-llama/Llama-4-Scout-17B-16E-Instruct/blob/main/LICENSE}}
    vLLM~\cite{kwon2023efficient}, used to serve both models, is released
    under the Apache License 2.0.\footnote{\url{https://github.com/vllm-project/vllm/blob/main/LICENSE}}

    \item \textbf{Hosted LLM APIs.}
    Gemini~\cite{gemini} and GPT~\cite{openai2026gpt56} are proprietary hosted services
    used under the Gemini API Terms and the OpenAI Services Agreement,
    respectively.\footnote{\url{https://ai.google.dev/gemini-api/terms}}
    \footnote{\url{https://openai.com/policies/services-agreement/}}
    We do not redistribute either model's weights.
\end{itemize}

We will release only our independently created annotation layer and code, which enable licensed users to combine the annotations with their own copies of MADAR and MADAR-CODA, subject to the original resource licenses.

\section{Model Sizes, Hyperparameters, and Computational Budget}
\label{app:compute}

This section reports model size, the training and inference hyperparameters, and the computational budget of the parsing and alignment experiments.
Table~\ref{tab:model-sizes} distinguishes locally available models from proprietary APIs, for which parameter counts are not disclosed.

\begin{table*}[htbp]
\centering
\small
\setlength{\tabcolsep}{5pt}
\begin{tabular}{@{}p{4.2cm}rp{8.2cm}@{}}
\toprule
\textbf{System} & \textbf{Parameters} & \textbf{Role and access} \\
\midrule
CAMeLBERT-Mix biaffine parser & 110,936,282
& Locally trained supervised parser used for the reported All Treebanks
results; the count includes the CAMeLBERT-Mix encoder and SuPar parsing head. \\
CAMeLBERT-Mix (SimAlign) & 109.1M
& Frozen pretrained encoder used to obtain contextual token representations;
SimAlign requires no task-specific training. \\
CAMeLBERT-Mix (Awesome-Align) & 132.7M
& Locally fine-tuned masked-language-model checkpoint, including its
alignment-training heads. \\
Fanar-2-27B-Instruct & 27B
& Open-weight model served locally with vLLM. \\
Llama-4-Scout & 109B total; 17B active
& Open-weight mixture-of-experts model served locally with vLLM. \\
Gemini; GPT & Not disclosed
& Proprietary models accessed through their providers' Batch APIs. \\
\bottomrule
\end{tabular}
\caption{Model sizes and access methods.}
\label{tab:model-sizes}
\end{table*}

\paragraph{Supervised Parsing}
Table~\ref{tab:parser-hparams} gives the fixed configuration used to train the
reported supervised parsers.
Each model was selected by development LAS.
The All Treebanks model used the 53,037-sentence pool described in Section~\ref{sec:approach}; the other newly trained models used the same configuration with their respective training subsets.
Multi-root decoding is an inference constraint and does not add model parameters.

\begin{table}[htbp]
\centering
\small
\setlength{\tabcolsep}{4pt}
\begin{tabular}{@{}lr@{\hspace{1.2em}}lr@{}}
\toprule
\textbf{Parameter} & \textbf{Value} & \textbf{Parameter} & \textbf{Value} \\
\midrule
Parser & Biaffine & Epochs & 20 \\
BERT layers mixed & 4 & BERT pooling & Mean \\
Arc MLP & 500 & Relation MLP & 100 \\
Encoder dropout & 0.1 & MLP dropout & 0.33 \\
Learning rate & $5\times10^{-5}$ & BERT LR rate & 10 \\
Batch token budget & 5,000 & Update steps & 4 \\
Warmup proportion & 0.1 & Gradient clip & 5.0 \\
$\beta_1,\beta_2$ & 0.9, 0.999 & $\epsilon$ & $10^{-8}$ \\
Weight decay & 0 & Seed & 1 \\
Decoder & \multicolumn{3}{l}{Projective multi-root} \\
\bottomrule
\end{tabular}
\caption{Hyperparameters for the reported CAMeLBERT-Mix supervised parsers.}
\label{tab:parser-hparams}
\end{table}

\paragraph{LLM Inference}
Each full LLM run contains 3,600 requests: six varieties, two splits, and both
zero- and 10-shot conditions, with 150 sentences per condition.
Fanar and Llama used bfloat16 vLLM inference on four A100 80GB GPUs with tensor-parallel
size 4, a 32,768-token context, a maximum of 8,192 generated tokens,
temperature 0, prefix caching, and XGrammar structured decoding.
The scheduler allowed at most 128 sequences and 32,768 batched tokens.
Gemini used its Batch API with a structured response schema and an 8,192-token output limit.
GPT used the Responses Batch API with structured output.

\paragraph{Alignment Experiments}
Table~\ref{tab:alignment-hparams} records both training and extraction
settings.
The training and inference pool contains all 21 unordered pairs induced by the six varieties, including six self-pairs, for 21,000 sentence pairs in total.
Evaluation in Section~\ref{sec:approach} uses the 15 cross-variety pairs.

\begin{table*}[htbp]
\centering
\small
\setlength{\tabcolsep}{5pt}
\begin{tabular}{@{}p{3.2cm}p{12.4cm}@{}}
\toprule
\textbf{System/stage} & \textbf{Configuration} \\
\midrule
SimAlign
& Frozen CAMeLBERT-Mix encoder; BPE tokenization; layer 8; intersection of
the bidirectional argmax alignments. No alignment-model training. \\
Awesome-Align training
& CAMeLBERT-Mix initialization; MLM, TLM, full bidirectional TLM, SO, and PSI
objectives; contrastive objective disabled; 5 epochs (1,645 optimizer updates);
batch size 64; gradient accumulation 1; maximum block length 128; learning
rate $2\times10^{-5}$; weight decay 0.01; 165 warmup updates; maximum gradient
norm 1.0; seed 42; one A100 GPU with TF32. \\
Awesome-Align extraction
& Layer 8; bidirectional softmax intersection; threshold 0.001; batch size
512; 8 data workers. \\
MGIZA++ training
& Both directions for every pair; 5 IBM Model~1, 5 HMM, 3 Model~3, and 3
Model~4 EM iterations; Models~2 and 5 disabled; 50 word classes; 4 threads per
direction and 8 concurrent pair jobs. Directional links were combined with
grow-diag-final-and. \\
\bottomrule
\end{tabular}
\caption{Training and extraction settings for the alignment experiments.}
\label{tab:alignment-hparams}
\end{table*}

\paragraph{Runtime and Cost}
Training the 20-epoch All Treebanks parser on 53,037 sentences took 36.1 minutes on one A100 80GB GPU.
The 11 additional supervised parser candidates in the optimization sweep took 828.9 minutes cumulatively. Thus, the 12 supervised models with preserved training logs required 865.0 minutes
of cumulative single-GPU time.
These measurements cover successful training, including epoch-level development evaluation and checkpoint selection, but exclude scheduler waiting and failed or preempted work.

Table~\ref{tab:compute-budget} reports inference time on the evaluated data.
For hosted models, time is the observed API-plus-evaluation turnaround and cost is reconstructed from preserved token usage. We do not estimate a monetary cost for local GPU use.

\begin{table*}[htbp]
\centering
\small
\setlength{\tabcolsep}{6pt}
\begin{tabular}{@{}lrlr@{}}
\toprule
\textbf{System} & \textbf{Time (min)}
& \textbf{Hardware/access} & \textbf{API cost (USD)} \\
\midrule
Gemini-3.6-Flash & 41.7 & Hosted Batch API & \$30.92 \\
GPT-5.6-Sol & 29.3 & Hosted Batch API & \$35.31 \\
Fanar-2-27B-Instruct & 14.0 & 4 $\times$ A100 80GB & -- \\
Llama-4-Scout & 27.4 & 4 $\times$ A100 80GB & -- \\
All Treebanks parser & 0.21 & 1 $\times$ A100 80GB & -- \\
\bottomrule
\end{tabular}
\caption{Inference runtime on the evaluated data and hosted API cost. The parser
measurement uses projective multi-root decoding and excludes model loading.}
\label{tab:compute-budget}
\end{table*}

The two hosted runs together cost \$66.23 and required 70.9 minutes of summed
turnaround.
Fanar processed the same workload in about half the clean model time of Llama (14.0 versus 27.4 minutes). The supervised parser was substantially faster at inference, processing the matched evaluation workload
in 0.21 minutes after warm-up.

\section{Full Test-Set Parsing Results}
\label{app:parsing-results}

Table~\ref{tab:appendix-parsing-test} reports the complete test matrix summarized by the five development-selected configurations in Table~\ref{tab:parsing-test}.

\begin{table*}[htbp]
\centering
\small
\setlength{\tabcolsep}{4pt}
\begin{tabular}{llcccccccc}
\toprule
\textbf{Model}
& \textbf{Training Data / Prompting}\phantom{XXXXXXXX}
& \textbf{Beirut}
& \textbf{Cairo}
& \textbf{Doha}
& \textbf{Rabat}
& \textbf{Tunis}
& \textbf{MSA}
& \textbf{Dialects}
& \textbf{All} \\
\midrule

\textbf{Supervised}
& {\cmtb} Beirut
& 79.6 & 76.1 & 80.0 & 69.3 & 75.1 & 69.8 & 75.9 & 74.8 \\

\textbf{Supervised}
& {\cmtb} Cairo
& 70.9 & 80.8 & 78.7 & 66.2 & 77.3 & 72.4 & 74.7 & 74.3 \\

\textbf{Supervised}
& {\cmtb} Doha
& 71.2 & 77.0 & 82.9 & 66.9 & 75.8 & 74.3 & 74.5 & 74.5 \\

\textbf{Supervised}
& {\cmtb} Rabat
& 63.7 & 72.1 & 73.1 & 79.4 & 76.3 & 70.2 & 73.0 & 72.5 \\

\textbf{Supervised}
& {\cmtb} Tunis
& 64.5 & 72.3 & 74.9 & 69.6 & 83.0 & 71.5 & 72.8 & 72.6 \\

\textbf{Supervised}
& {\cmtb} MSA
& 59.8 & 68.9 & 69.4 & 62.9 & 68.0 & 82.5 & 65.7 & 68.7 \\

\midrule
\textbf{Supervised}
& {\cmtb} Dialects
& 83.6 & 84.2 & 88.7 & 84.8 & 86.3 & 78.5 & 85.4 & 84.2 \\

\textbf{Supervised}
& {\cmtb} All
& 84.8 & 84.5 & 87.8 & 82.7 & 86.0 & 85.1 & 85.1 & 85.1 \\

\midrule
\textbf{Supervised}
& PATB + CamelTB
& 70.9 & 76.1 & 80.2 & 69.7 & 75.3 & 90.3 & 74.3 & 77.2 \\

\textbf{Supervised}
& PATB + CamelTB + ARZTB + Gumar
& 77.8
& 86.1
& 85.5
& 74.9
& 79.6
& \textbf{90.4}
& 80.7
& 82.4 \\

\midrule
\textbf{Supervised}
& All Treebanks
& \textbf{86.9}
& \textbf{87.4}
& \textbf{90.7}
& \textbf{85.4}
& 87.9
& 86.9
& \textbf{87.6}
& \textbf{87.5} \\

\midrule
\textbf{Gemini}
& 0-shot
& 69.8 & 71.7 & 71.2 & 68.2 & 73.2 & 75.9 & 70.8 & 71.7 \\

\textbf{Gemini}
& 10-shot
& \underline{85.8}
& \underline{83.3}
& \underline{86.9}
& \underline{83.8}
& \textbf{\underline{88.1}}
& \underline{86.4}
& \underline{85.5}
& \underline{85.7} \\

\midrule
\textbf{GPT}
& 0-shot
& 63.7 & 69.4 & 63.8 & 63.9 & 64.4 & 71.3 & 65.1 & 66.2 \\

\textbf{GPT}
& 10-shot
& 84.5 & 81.9 & 84.5 & 82.4 & 85.0 & 85.1 & 83.6 & 83.9 \\

\midrule
\textbf{Fanar}
& 0-shot
& 13.5 & 15.4 & 16.0 & 17.3 & 17.1 & 17.6 & 15.9 & 16.2 \\

\textbf{Fanar}
& 10-shot
& 67.6 & 70.0 & 72.3 & 72.6 & 72.5 & 73.3 & 71.0 & 71.4 \\

\midrule
\textbf{Llama}
& 0-shot
& 22.9 & 23.8 & 24.9 & 25.0 & 22.6 & 31.5 & 23.8 & 25.2 \\

\textbf{Llama}
& 10-shot
& 63.4 & 66.9 & 70.2 & 67.5 & 69.0 & 69.4 & 67.3 & 67.7 \\

\bottomrule
\end{tabular}
\caption{
test-set parsing results (LAS) for supervised parsers and prompted
LLMs.
All \textbf{Supervised} rows use the CAMeLBERT-Mix Biaffine  model and projective
multi-root decoding; their configurations differ only in training data.
All Treebanks denotes PATB, CamelTB, ARZTB, Gumar, and {\cmtb}.
Bold indicates the best result overall, and underlining indicates the best LLM
result.
}
\label{tab:appendix-parsing-test}
\end{table*}

\paragraph{Paired Bootstrap Confidence Intervals}
Table~\ref{tab:bootstrap-parsing} reports paired bootstrap intervals for the development- and test-set claims discussed in Section~\ref{sec:eval}. Within each split, a resampled cluster is one underlying sentence ID together with all five dialect realizations for the Dialects aggregate or all six realizations for the All aggregate.
The intervals therefore preserve the dependence created by the parallel design.
Twelve of fourteen development intervals and thirteen of fourteen test intervals exclude zero.
On development data, the exceptions are adding MSA to the pooled dialect data and Gemini versus {\cmtb} All; on test data, only Gemini versus {\cmtb} All includes zero.
The intervals quantify uncertainty over evaluation sentences conditional on the saved system outputs and do not include variation from retraining the supervised models.

\begin{table*}[htbp]
\centering
\small
\setlength{\tabcolsep}{5pt}
\begin{tabular}{@{}lllcrr@{}}
\toprule
\textbf{Claim} & \textbf{System A} & \textbf{System B} &
\textbf{Set} & \textbf{$\Delta$ LAS} & \textbf{95\% CI} \\
\midrule
\multicolumn{6}{@{}l}{\textit{Development set}} \\
\addlinespace[2pt]
Dialect pooling & {\cmtb} Dialects & {\cmtb} Beirut & Dialects & 9.9 & [8.6, 11.2] \\
& {\cmtb} Dialects & {\cmtb} Cairo & Dialects & 11.1 & [9.8, 12.4] \\
& {\cmtb} Dialects & {\cmtb} Doha & Dialects & 10.3 & [9.0, 11.6] \\
& {\cmtb} Dialects & {\cmtb} Rabat & Dialects & 12.0 & [10.5, 13.5] \\
& {\cmtb} Dialects & {\cmtb} Tunis & Dialects & 12.7 & [11.2, 14.3] \\
\midrule
Training coverage & {\cmtb} All & {\cmtb} Dialects & All & 0.5 & [$-$0.1, 1.1] \\
& All Treebanks & Four external treebanks & All & 5.2 & [4.1, 6.2] \\
& All Treebanks & {\cmtb} All & All & 2.6 & [1.6, 3.6] \\
\midrule
Retrieved examples & Gemini 10-shot & Gemini 0-shot & All & 15.3 & [13.8, 17.0] \\
& GPT 10-shot & GPT 0-shot & All & 18.1 & [16.0, 20.3] \\
& Fanar 10-shot & Fanar 0-shot & All & 58.8 & [56.4, 61.3] \\
& Llama 10-shot & Llama 0-shot & All & 47.6 & [45.3, 50.0] \\
\midrule
Best systems & All Treebanks & Gemini 10-shot & All & 1.6 & [0.5, 2.7] \\
& Gemini 10-shot & {\cmtb} All & All & 1.0 & [$-$0.2, 2.3] \\
\midrule\midrule
\multicolumn{6}{@{}l}{\textit{Test set}} \\
\addlinespace[2pt]
Dialect pooling & {\cmtb} Dialects & {\cmtb} Beirut & Dialects & 9.5 & [8.3, 10.8] \\
& {\cmtb} Dialects & {\cmtb} Cairo & Dialects & 10.8 & [9.2, 12.3] \\
& {\cmtb} Dialects & {\cmtb} Doha & Dialects & 10.9 & [9.4, 12.4] \\
& {\cmtb} Dialects & {\cmtb} Rabat & Dialects & 12.5 & [10.8, 14.1] \\
& {\cmtb} Dialects & {\cmtb} Tunis & Dialects & 12.7 & [11.0, 14.4] \\
\midrule
Training coverage & {\cmtb} All & {\cmtb} Dialects & All & 0.9 & [0.2, 1.6] \\
& All Treebanks & Four external treebanks & All & 5.0 & [3.7, 6.5] \\
& All Treebanks & {\cmtb} All & All & 2.4 & [1.3, 3.4] \\
\midrule
Retrieved examples & Gemini 10-shot & Gemini 0-shot & All & 14.0 & [12.6, 15.5] \\
& GPT 10-shot & GPT 0-shot & All & 17.7 & [15.9, 19.5] \\
& Fanar 10-shot & Fanar 0-shot & All & 55.2 & [52.6, 58.0] \\
& Llama 10-shot & Llama 0-shot & All & 42.5 & [40.0, 45.3] \\
\midrule
Best systems & All Treebanks & Gemini 10-shot & All & 1.8 & [0.6, 3.0] \\
& Gemini 10-shot & {\cmtb} All & All & 0.6 & [$-$0.5, 1.7] \\
\bottomrule
\end{tabular}
\caption{Paired sentence-ID-clustered percentile bootstrap intervals for
principal development- and test-set LAS comparisons (10,000 replicates per
comparison). $\Delta$ LAS is System A minus System B in absolute percentage
points. Within each split, every replicate resamples all 150 underlying
sentence IDs with replacement and retains every relevant parallel variety
realization.}
\label{tab:bootstrap-parsing}
\end{table*}

\paragraph{Performance on Established Treebanks}
Table~\ref{tab:external-treebank-test} compares our All Treebanks model with
CamelParser2.0~\citep{elshabrawy-etal-2023-camelparser2} and
CAMeLParser-Dialects~\citep{faroukzakariaelshabrawy-etal-2026-parsing} on the test
sets of four established treebanks. Our model achieves the highest LAS on
Gumar, ARZTB, and CamelTB, improving over the strongest previous result by
7.7, 0.5, and 1.7 points, respectively. It also remains within 0.1 points of
CamelParser2.0 on PATB. These results establish new state-of-the-art
performance on three external treebanks, showing that the gains extend beyond
the proposed {\cmtb} benchmark.

\begin{table*}[htbp]
\centering
\footnotesize
\setlength{\tabcolsep}{3pt}
\begin{tabular}{@{}lp{3.9cm}lrrrrr@{}}
\toprule
\textbf{Model} & \textbf{Training Data} & \textbf{Decoding}
& \textbf{Gumar} & \textbf{ARZTB} & \textbf{CamelTB}
& \textbf{PATB} & \textbf{{\cmtb} All}  \\
\midrule

CamelParser2.0
& PATB + CamelTB
& Single-root
& 72.7 & 73.2 & 87.5 & \textbf{91.3} & 76.0 \\

CAMeLParser-Dialects
& PATB + CamelTB + ARZTB + Gumar
& Single-root
& 80.3 & 84.2 & 87.2 & 90.8 & 81.2 \\

\textbf{Ours}
& All Treebanks
& Single-root
& 80.4 & 84.1 & 87.5 & 90.7 & 86.5 \\

\textbf{Ours}
& All Treebanks
& Multi-root
& \textbf{88.0} & \textbf{84.7} & \textbf{89.2}
& 91.2 & \textbf{87.5} \\

\bottomrule
\end{tabular}
\caption{Test-set LAS on established Arabic dependency treebanks and {\cmtb}. The published baselines are
CamelParser2.0~\citep{elshabrawy-etal-2023-camelparser2} and
CAMeLParser-Dialects~\citep{faroukzakariaelshabrawy-etal-2026-parsing}.
``All Treebanks'' denotes PATB, CamelTB, ARZTB, Gumar, and {\cmtb}. Bold
indicates the best result for each test set.}
\label{tab:external-treebank-test}
\end{table*}
\section{Additional Corpus Statistics}
\label{app:corpus-statistics}

We report lexical, structural, POS, and dependency-relation statistics for the six parallel treebanks.
Table~\ref{tab:treebank-descriptive-statistics} summarizes the lexical and structural characteristics of the six parallel treebanks.
Despite their shared content, the varieties range from 7,768 tokens in Doha to 9,294 in MSA.
MSA has the longest sentences and deepest trees on average, while Doha has the shortest and shallowest.
Mean dependency distance remains similar across varieties (1.94--2.02 tokens), while 10.9--18.1\% of sentences contain multiple roots.

Table~\ref{tab:treebank-pos} presents POS distributions across the six varieties.
The POS profiles are also similar: nominals constitute 50.7\% of all tokens, followed by punctuation (17.3\%), particles (17.2\%), and active verbs (12.7\%).
MSA has more particles and is the only variety with passive verbs, whereas Doha and Tunis have more active verbs.

Table~\ref{tab:treebank-relations} summarizes the distribution of dependency relations across the six varieties.
Five dependency relations account for 98.5\% of tokens: \textit{MOD} (45.1\%), \textit{OBJ} (21.1\%), \textit{---} (14.7\%), \textit{IDF} (10.2\%), and \textit{SBJ} (7.4\%).
The largest variation is in \textit{MOD} (42.9--49.5\%), while MSA uses \textit{PRD} more often than the dialects.

\begin{table*}[htbp]
\centering
\small
\setlength{\tabcolsep}{4pt}
\begin{tabular}{lrrrrrrrr}
\toprule
& \multicolumn{2}{c}{\textbf{Corpus size}}
& \multicolumn{2}{c}{\textbf{Lexicon and length}}
& \multicolumn{2}{c}{\textbf{Tree structure}}
& \multicolumn{2}{c}{\textbf{Roots}} \\
\cmidrule(lr){2-3} \cmidrule(lr){4-5} \cmidrule(lr){6-7} \cmidrule(lr){8-9}
\textbf{Variety} & \textbf{Sent.} & \textbf{Tokens} & \textbf{Unique} &
\textbf{Sent. len.} & \textbf{Max depth} & \textbf{Dep. dist.} &
\textbf{Roots/sent.} & \textbf{Multi-root (\%)} \\
\midrule
\textbf{Beirut} & 1,000 & 8,462 & 1,948 & 8.46 & 3.03 & 1.97 & 1.21 & 16.9 \\
\textbf{Cairo} & 1,000 & 8,643 & 2,005 & 8.64 & 3.27 & 2.01 & 1.14 & 10.9 \\
\textbf{Doha} & 1,000 & 7,768 & 1,846 & 7.77 & 2.97 & 1.94 & 1.20 & 16.7 \\
\textbf{Rabat} & 1,000 & 8,925 & 2,089 & 8.93 & 3.26 & 2.02 & 1.19 & 15.7 \\
\textbf{Tunis} & 1,000 & 8,469 & 2,020 & 8.47 & 3.21 & 2.01 & 1.19 & 16.3 \\
\textbf{MSA} & 1,000 & 9,294 & 2,314 & 9.29 & 3.65 & 1.96 & 1.22 & 18.1 \\
\bottomrule
\end{tabular}
\caption{Descriptive statistics for the six parallel treebanks. Unique
forms are distinct token forms. Mean maximum depth assigns roots depth
zero. Mean dependency distance is the absolute difference between the
token positions of a non-root dependent and its head, averaged first
within each sentence and then across sentences. Roots/sent. is the mean
number of tokens with \texttt{HEAD=0} per sentence; multi-root is the
percentage of sentences containing more than one such token.}
\label{tab:treebank-descriptive-statistics}
\end{table*}

\begin{table*}[htbp]
\centering
\small
\setlength{\tabcolsep}{8pt}
\begin{tabular}{lrrrrrrr}
\toprule
\textbf{POS} & \textbf{Beirut} & \textbf{Cairo} & \textbf{Doha} &
\textbf{Rabat} & \textbf{Tunis} & \textbf{MSA} & \textbf{All} \\
\midrule
NOM & 50.5 & 53.3 & 49.9 & 50.8 & 50.2 & 49.2 & 50.7 \\
PNX & 17.1 & 17.3 & 19.2 & 17.0 & 17.7 & 15.6 & 17.3 \\
PRT & 19.0 & 16.2 & 13.8 & 16.7 & 14.5 & 22.1 & 17.2 \\
VRB & 11.2 & 10.8 & 14.5 & 13.4 & 15.4 & 11.0 & 12.7 \\
PROP & 2.2 & 2.4 & 2.6 & 2.0 & 2.1 & 1.7 & 2.1 \\
VRB-PASS & 0.0 & 0.0 & 0.0 & 0.0 & 0.0 & 0.5 & 0.1 \\
\bottomrule
\end{tabular}
\caption{POS distributions (\%) over all data splits.}
\label{tab:treebank-pos}
\end{table*}

\begin{table*}[htbp]
\centering
\small
\setlength{\tabcolsep}{8pt}
\begin{tabular}{lrrrrrrr}
\toprule
\textbf{Relation} & \textbf{Beirut} & \textbf{Cairo} & \textbf{Doha} &
\textbf{Rabat} & \textbf{Tunis} & \textbf{MSA} & \textbf{All} \\
\midrule
MOD & 45.1 & 44.4 & 42.9 & 49.5 & 44.7 & 43.8 & 45.1 \\
OBJ & 21.4 & 19.8 & 21.3 & 19.9 & 22.9 & 21.3 & 21.1 \\
--- & 15.1 & 13.9 & 16.5 & 14.2 & 14.8 & 13.8 & 14.7 \\
IDF & 10.3 & 11.3 & 9.5 & 8.7 & 9.9 & 11.4 & 10.2 \\
SBJ & 6.9 & 8.9 & 8.5 & 6.4 & 6.6 & 7.4 & 7.4 \\
PRD & 0.8 & 1.3 & 0.9 & 0.7 & 0.8 & 1.9 & 1.1 \\
TPC & 0.1 & 0.2 & 0.2 & 0.3 & 0.2 & 0.2 & 0.2 \\
TMZ & 0.2 & 0.2 & 0.2 & 0.2 & 0.2 & 0.2 & 0.2 \\
\bottomrule
\end{tabular}
\caption{Dependency-relation distributions (\%) over all data splits.}
\label{tab:treebank-relations}
\end{table*}

\section{Pairwise Alignment Results}
\label{app:syntax-similarity}

Table~\ref{tab:appendix-alignment} reports the alignment-level measures
for every analyzed unordered variety pair.  Table~\ref{tab:manual-alignment-evaluation}
describes the complete manually annotated evaluation data and reports
the one-to-one aligner results.

\begin{table*}[htbp]
\centering
\small
\setlength{\tabcolsep}{4pt}
\begin{tabular}{@{}llrrrrrrr@{}}
\toprule
\multicolumn{2}{c}{\textbf{Variety pair}} &
\multicolumn{1}{c}{\textbf{All tokens}} &
\multicolumn{2}{c}{\textbf{Aligned token pairs}} &
\multicolumn{4}{c}{\textbf{Head-aligned token pairs}} \\
\cmidrule(lr){1-2}
\cmidrule(lr){3-3}
\cmidrule(lr){4-5}
\cmidrule(lr){6-9}
\multicolumn{2}{c}{\textbf{Pair}} &
\textbf{Aligned} &
\textbf{Token POS} &
\textbf{Head also} &
\textbf{Token+head POS} &
\textbf{Relation} &
\textbf{Direction} &
\textbf{Complete} \\
& &
\textbf{(\%)} &
\textbf{match (\%)} &
\textbf{aligned (\%)} &
\textbf{match (\%)} &
\textbf{match (\%)} &
\textbf{match (\%)} &
\textbf{match (\%)} \\
\midrule
Beirut & Cairo & 85.6 & 90.8 & 67.1 & 86.4 & 93.4 & 98.3 & 81.6 \\
Beirut & Doha  & 87.3 & 89.7 & 71.4 & 79.2 & 93.9 & 99.2 & 76.0 \\
Beirut & MSA   & 82.5 & 87.3 & 58.6 & 75.3 & 90.7 & 99.1 & 70.7 \\
Beirut & Rabat & 82.8 & 88.0 & 65.6 & 74.6 & 92.6 & 99.0 & 70.8 \\
Beirut & Tunis & 83.0 & 87.3 & 66.9 & 74.6 & 92.1 & 98.6 & 70.1 \\
Cairo & Doha   & 86.6 & 91.4 & 72.6 & 82.2 & 94.1 & 98.5 & 78.1 \\
Cairo & MSA    & 82.4 & 86.7 & 58.5 & 75.5 & 90.0 & 98.1 & 70.1 \\
Cairo & Rabat  & 82.1 & 87.6 & 62.5 & 72.3 & 91.0 & 97.7 & 66.7 \\
Cairo & Tunis  & 83.5 & 86.7 & 66.8 & 70.8 & 91.0 & 99.0 & 66.8 \\
Doha & MSA     & 82.5 & 87.6 & 58.7 & 82.5 & 90.0 & 99.0 & 77.9 \\
Doha & Rabat   & 83.5 & 88.9 & 67.3 & 79.9 & 91.4 & 99.1 & 75.6 \\
Doha & Tunis   & 84.8 & 87.7 & 69.6 & 77.8 & 91.8 & 98.7 & 73.7 \\
MSA & Rabat    & 80.3 & 87.6 & 57.0 & 79.9 & 89.5 & 98.9 & 74.8 \\
MSA & Tunis    & 79.9 & 86.8 & 56.9 & 80.1 & 90.1 & 98.4 & 74.6 \\
Rabat & Tunis  & 83.3 & 92.8 & 68.4 & 89.5 & 93.2 & 98.3 & 83.8 \\
\midrule
\multicolumn{2}{l}{\textbf{Pair average}} & 83.3 & 88.5 & 64.5 & 78.7 & 91.6 & 98.7 & 74.1 \\
\bottomrule
\end{tabular}
\caption{Alignment-level structural agreement from the selected
Awesome-Align links by unordered variety pair; all values are percentages.
The columns follow the successive evaluation subsets shown in the header.
Aligned is the mean
of the two varieties' token-coverage percentages.  For aligned token
pairs, we report token POS match and the percentage whose heads also
align. Both POS requires agreement in token and head POS. The final four
measures are evaluated on this head-aligned subset; complete match requires
both POS, relation, and
direction to all match.}
\label{tab:appendix-alignment}
\end{table*}

\begin{table*}[htbp]
\centering
\small
\setlength{\tabcolsep}{10pt}
\begin{tabular}{lrrrr}
\toprule
\textbf{Manual evaluation data} & \textbf{Sentence pairs} &
\textbf{Source tokens} & \textbf{Target tokens} &
\textbf{Manual links} \\
\midrule
Complete annotated sample & 300 & 2,545 & 2,615 & 2,172 \\
\midrule
\textbf{Aligner} & \textbf{Predicted links} & \textbf{Precision} &
\textbf{Recall} & \textbf{F$_1$} \\
\midrule
SimAlign & 2,036 & \textbf{91.9} & 86.1 & 88.9 \\
Awesome-Align & 2,227 & 91.1 & \textbf{93.4} & \textbf{92.2} \\
MGIZA++ & 2,215 & 79.0 & 80.6 & 79.8 \\
\bottomrule
\end{tabular}
\caption{Manual one-to-one alignment data and system evaluation.  The
upper block describes the complete sample annotated by two annotators;
the lower block reports each system's predicted links and
micro-averaged precision, recall, and F$_1$.  These metrics compare each
predicted link set with the manually annotated non-null links.}
\label{tab:manual-alignment-evaluation}
\end{table*}

\end{document}

%% file: macros.tex
\newcommand{\hide}[1]{}

\newcommand{\AYN}{{$\varsigma$}}

%% file: camel-bib-v3.bib
@misc{Buckwalter:2002:buckwalter,
	Author = {Tim Buckwalter},
	Howpublished = {Linguistic Data Consortium (LDC) catalog number LDC2002L49, ISBN 1-58563-257-0},
	Title = {Buckwalter {{A}rabic} Morphological Analyzer Version 1.0},
	Year = {2002}}

@article{Ferguson:1959:diglossia,
	Author = {Charles F Ferguson},
	Journal = {Word},
	Number = {2},
	Pages = {325--340},
	Title = {{Diglossia}},
	Volume = {15},
	Year = {1959}}

@inproceedings{Gao:2008:parallel,
	Address = {Columbus, Ohio},
	Author = {Gao, Qin and Vogel, Stephan},
	Booktitle = {Software Engineering, Testing, and Quality Assurance for Natural Language Processing},
	Pages = {49--57},
	Title = {Parallel implementations of word alignment tool},
	Year = {2008}}

@inproceedings{Habash:2009:catib,
	Address = {Suntec, Singapore},
	Author = {Habash, Nizar and Roth, Ryan},
	Booktitle = {Proceedings of the Joint Conference of the Association for Computational Linguistics and the International Joint Conference on Natural Language Processing (ACL-IJCNLP)},
	Pages = {221--224},
	Title = {{CATiB: The Columbia {A}rabic Treebank}},
	Year = {2009}}

@inproceedings{Habash:2009:syntactic,
	Address = {Cairo, Egypt},
	Author = {Nizar Habash and Reem Faraj and Ryan Roth},
	Booktitle = {Proceedings of the International Conference on {A}rabic Language Resources and Tools},
	Title = {{Syntactic Annotation in the Columbia {A}rabic Treebank}},
	Year = {2009}}

@inproceedings{Hajic:2004:prague,
	Address = {Cairo, Egypt},
	Author = {Jan Haji{\v{c}} and Otakar Smr{\v{z}} and Petr Zem{\'{a}}nek and Jan {\v{S}}naidauf and Emanuel Be{\v{s}}ka},
	Booktitle = {Proceedings of the International Conference on {A}rabic Language Resources and Tools},
	Pages = {110--117},
	Publisher = {ELDA},
	Title = {{Prague {A}rabic Dependency Treebank: Development in Data and Tools}},
	Year = {2004}}

@inproceedings{Khalifa:2018:morphologically,
	Address = {Miyazaki, Japan},
	Author = {Salam Khalifa and Nizar Habash and Fadhl Eryani and Ossama Obeid and Dana Abdulrahim and Meera Al Kaabi},
	Booktitle = {Proceedings of the Language Resources and Evaluation Conference (LREC)},
	Title = {A Morphologically Annotated Corpus of {E}mirati {A}rabic},
	Year = {2018}}

@inproceedings{Maamouri:2004:penn,
	Address = {Cairo, Egypt},
	Author = {Mohamed Maamouri and Ann Bies and Tim Buckwalter and Wigdan Mekki},
	Booktitle = {Proceedings of the International Conference on {A}rabic Language Resources and Tools},
	Pages = {102--109},
	Title = {{The Penn {{A}rabic} Treebank: Building a Large-Scale Annotated {{A}rabic} Corpus}},
	Year = {2004}}

@inproceedings{Maamouri:2008:enhancing,
	Address = {Marrakech, Morocco},
	Author = {Mohamed Maamouri and Ann Bies and Seth Kulick},
	Booktitle = {Proceedings of the Language Resources and Evaluation Conference (LREC)},
	Title = {Enhancing the {A}rabic Treebank: a Collaborative Effort toward New Annotation Guidelines},
	Year = {2008}}

@misc{Maamouri:2012:arz-4,
	Author = {Mohamed Maamouri and Ann Bies and Seth Kulick and Sondos Krouna and Dalila Tabassi and Michael Ciul},
	Howpublished = {LDC catalog number LDC2012E99},
	Title = {{Egyptian {A}rabic Treebank DF Part 4 V2.0}},
	Year = {2012}}

@article{Marton:2013:dependency,
	Author = {Marton, Yuval and Habash, Nizar and Rambow, Owen},
	Journal = {Computational Linguistics},
	Number = {1},
	Pages = {161--194},
	Title = {Dependency parsing of Modern Standard {{A}rabic} with lexical and inflectional features},
	Volume = {39},
	Year = {2013}}

@inproceedings{Shahrour:2016:camelparser,
	Author = {Shahrour, Anas and Khalifa, Salam and Taji, Dima and Habash, Nizar},
	Booktitle = {Proceedings of the International Conference on Computational Linguistics (COLING)},
	Pages = {228--232},
	Title = {{CamelParser: A system for {A}rabic syntactic analysis and morphological disambiguation}},
	Year = {2016}}

@inproceedings{Taji:2018:arabic,
	Address = {Miyazaki, Japan},
	Author = {Taji, Dima and El Gizuli, Jamila and Habash, Nizar},
	Booktitle = {Proceedings of the Workshop on Open-Source {A}rabic Corpora and Processing Tools (OSACT)},
	Title = {An {{A}rabic} Dependency Treebank in the Travel Domain},
	Year = {2018}}

@inproceedings{mokh-etal-2024-domain,
    title = "Out-of-{D}omain {D}ependency {P}arsing for {D}ialects of {A}rabic: {A} {C}ase {S}tudy",
    author = {Abo Mokh, Noor  and
      Dakota, Daniel  and
      K{\"u}bler, Sandra},
    editor = "Habash, Nizar  and
      Bouamor, Houda  and
      Eskander, Ramy  and
      Tomeh, Nadi  and
      Abu Farha, Ibrahim  and
      Abdelali, Ahmed  and
      Touileb, Samia  and
      Hamed, Injy  and
      Onaizan, Yaser  and
      Alhafni, Bashar  and
      Antoun, Wissam  and
      Khalifa, Salam  and
      Haddad, Hatem  and
      Zitouni, Imed  and
      AlKhamissi, Badr  and
      Almatham, Rawan  and
      Mrini, Khalil",
    booktitle = "Proceedings of the Second Arabic Natural Language Processing Conference",
    month = aug,
    year = "2024",
    address = "Bangkok, Thailand",
    publisher = "Association for Computational Linguistics",
    url = "https://aclanthology.org/2024.arabicnlp-1.16/",
    doi = "10.18653/v1/2024.arabicnlp-1.16",
    pages = "170--182"
}


%% file: custom.bib
@misc{diab2013ldcarabictreebanksassociated,
      title={{LDC} {Arabic} Treebanks and Associated Corpora: Data Divisions Manual}, 
      author={Mona Diab and Nizar Habash and Owen Rambow and Ryan Roth},
      year={2013},
      eprint={1309.5652},
      archivePrefix={arXiv},
      primaryClass={cs.CL},
      url={https://arxiv.org/abs/1309.5652}, 
}

@article{IriarteDiez2023whatiswhitearabic,
  author    = {Iriarte D{\'{i}}ez, Ana and Laaber, Claudia and Van Kampen, Nina and Ben{\'{i}}tez Fern{\'{a}}ndez, Montserrat},
  title     = {What is White {A}rabic? {N}ew Labels in a Changing {A}rab World},
  journal   = {Revista Espa{\~{n}}ola de Lingü{\"{u}}{\'{i}}stica},
  volume    = {53},
  number    = {2},
  pages     = {229--266},
  year      = {2023},
  doi       = {10.31810/rsel.53.2.9},
  url       = {https://doi.org/10.31810/rsel.53.2.9}
}

@inproceedings{shaban-habash-2025-arabic,
    title = "The {A}rabic Generality Score: Another Dimension of Modeling {A}rabic Dialectness",
    author = "Sha{'}ban, Sanad  and
      Habash, Nizar",
    editor = "Christodoulopoulos, Christos  and
      Chakraborty, Tanmoy  and
      Rose, Carolyn  and
      Peng, Violet",
    booktitle = "Proceedings of the 2025 Conference on Empirical Methods in Natural Language Processing",
    month = nov,
    year = "2025",
    address = "Suzhou, China",
    publisher = "Association for Computational Linguistics",
    url = "https://aclanthology.org/2025.emnlp-main.1524/",
    doi = "10.18653/v1/2025.emnlp-main.1524",
    pages = "30002--30013",
    ISBN = "979-8-89176-332-6"
}

@misc{gemini,
      title={Gemini: A Family of Highly Capable Multimodal Models}, 
      author={{Gemini Team} and Rohan Anil and Sebastian Borgeaud and Jean-Baptiste Alayrac and Jiahui Yu and Radu Soricut and Johan Schalkwyk and Andrew M. Dai and Anja Hauth and Katie Millican and David Silver and Melvin Johnson and Ioannis Antonoglou and Julian Schrittwieser and Amelia Glaese and Jilin Chen and Emily Pitler and Timothy Lillicrap and Angeliki Lazaridou and Orhan Firat and James Molloy and Michael Isard and Paul R. Barham and Tom Hennigan and Benjamin Lee and Fabio Viola and Malcolm Reynolds and Yuanzhong Xu and Ryan Doherty and Eli Collins and Clemens Meyer and Eliza Rutherford and Erica Moreira and Kareem Ayoub and Megha Goel and Jack Krawczyk and Cosmo Du and Ed Chi and Heng-Tze Cheng and Eric Ni and Purvi Shah and Patrick Kane and Betty Chan and Manaal Faruqui and Aliaksei Severyn and Hanzhao Lin and YaGuang Li and Yong Cheng and Abe Ittycheriah and Mahdis Mahdieh and Mia Chen and Pei Sun and Dustin Tran and Sumit Bagri and Balaji Lakshminarayanan and Jeremiah Liu and Andras Orban and Fabian Güra and Hao Zhou and Xinying Song and Aurelien Boffy and Harish Ganapathy and Steven Zheng and HyunJeong Choe and Ágoston Weisz and Tao Zhu and Yifeng Lu and Siddharth Gopal and Jarrod Kahn and Maciej Kula and Jeff Pitman and Rushin Shah and Emanuel Taropa and Majd Al Merey and Martin Baeuml and Zhifeng Chen and Laurent El Shafey and Yujing Zhang and Olcan Sercinoglu and George Tucker and Enrique Piqueras and Maxim Krikun and Iain Barr and Nikolay Savinov and Ivo Danihelka and Becca Roelofs and Anaïs White and Anders Andreassen and Tamara von Glehn and Lakshman Yagati and Mehran Kazemi and Lucas Gonzalez and Misha Khalman and Jakub Sygnowski and Alexandre Frechette and Charlotte Smith and Laura Culp and Lev Proleev and Yi Luan and Xi Chen and James Lottes and Nathan Schucher and Federico Lebron and Alban Rrustemi and Natalie Clay and Phil Crone and Tomas Kocisky and Jeffrey Zhao and Bartek Perz and Dian Yu and Heidi Howard and Adam Bloniarz and Jack W. Rae and Han Lu and Laurent Sifre and Marcello Maggioni and Fred Alcober and Dan Garrette and Megan Barnes and Shantanu Thakoor and Jacob Austin and Gabriel Barth-Maron and William Wong and Rishabh Joshi and Rahma Chaabouni and Deeni Fatiha and Arun Ahuja and Gaurav Singh Tomar and Evan Senter and Martin Chadwick and Ilya Kornakov and Nithya Attaluri and Iñaki Iturrate and Ruibo Liu and Yunxuan Li and Sarah Cogan and Jeremy Chen and Chao Jia and Chenjie Gu and Qiao Zhang and Jordan Grimstad and Ale Jakse Hartman and Xavier Garcia and Thanumalayan Sankaranarayana Pillai and Jacob Devlin and Michael Laskin and Diego de Las Casas and Dasha Valter and Connie Tao and Lorenzo Blanco and Adrià Puigdomènech Badia and David Reitter and Mianna Chen and Jenny Brennan and Clara Rivera and Sergey Brin and Shariq Iqbal and Gabriela Surita and Jane Labanowski and Abhi Rao and Stephanie Winkler and Emilio Parisotto and Yiming Gu and Kate Olszewska and Ravi Addanki and Antoine Miech and Annie Louis and Denis Teplyashin and Geoff Brown and Elliot Catt and Jan Balaguer and Jackie Xiang and Pidong Wang and Zoe Ashwood and Anton Briukhov and Albert Webson and Sanjay Ganapathy and Smit Sanghavi and Ajay Kannan and Ming-Wei Chang and Axel Stjerngren and Josip Djolonga and Yuting Sun and Ankur Bapna and Matthew Aitchison and Pedram Pejman and Henryk Michalewski and Tianhe Yu and Cindy Wang and Juliette Love and Junwhan Ahn and Dawn Bloxwich and Kehang Han and Peter Humphreys and Thibault Sellam and James Bradbury and Varun Godbole and Sina Samangooei and Bogdan Damoc and Alex Kaskasoli and Sébastien M. R. Arnold and Vijay Vasudevan and Shubham Agrawal and Jason Riesa and Dmitry Lepikhin and Richard Tanburn and Srivatsan Srinivasan and Hyeontaek Lim and Sarah Hodkinson and Pranav Shyam and Johan Ferret and Steven Hand and Ankush Garg and Tom Le Paine and Jian Li and Yujia Li and Minh Giang and Alexander Neitz and Zaheer Abbas and Sarah York and Machel Reid and Elizabeth Cole and Aakanksha Chowdhery and Dipanjan Das and Dominika Rogozińska and Vitaliy Nikolaev and Pablo Sprechmann and Zachary Nado and Lukas Zilka and Flavien Prost and Luheng He and Marianne Monteiro and Gaurav Mishra and Chris Welty and Josh Newlan and Dawei Jia and Miltiadis Allamanis and Clara Huiyi Hu and Raoul de Liedekerke and Justin Gilmer and Carl Saroufim and Shruti Rijhwani and Shaobo Hou and Disha Shrivastava and Anirudh Baddepudi and Alex Goldin and Adnan Ozturel and Albin Cassirer and Yunhan Xu and Daniel Sohn and Devendra Sachan and Reinald Kim Amplayo and Craig Swanson and Dessie Petrova and Shashi Narayan and Arthur Guez and Siddhartha Brahma and Jessica Landon and Miteyan Patel and Ruizhe Zhao and Kevin Villela and Luyu Wang and Wenhao Jia and Matthew Rahtz and Mai Giménez and Legg Yeung and James Keeling and Petko Georgiev and Diana Mincu and Boxi Wu and Salem Haykal and Rachel Saputro and Kiran Vodrahalli and James Qin and Zeynep Cankara and Abhanshu Sharma and Nick Fernando and Will Hawkins and Behnam Neyshabur and Solomon Kim and Adrian Hutter and Priyanka Agrawal and Alex Castro-Ros and George van den Driessche and Tao Wang and Fan Yang and Shuo-yiin Chang and Paul Komarek and Ross McIlroy and Mario Lučić and Guodong Zhang and Wael Farhan and Michael Sharman and Paul Natsev and Paul Michel and Yamini Bansal and Siyuan Qiao and Kris Cao and Siamak Shakeri and Christina Butterfield and Justin Chung and Paul Kishan Rubenstein and Shivani Agrawal and Arthur Mensch and Kedar Soparkar and Karel Lenc and Timothy Chung and Aedan Pope and Loren Maggiore and Jackie Kay and Priya Jhakra and Shibo Wang and Joshua Maynez and Mary Phuong and Taylor Tobin and Andrea Tacchetti and Maja Trebacz and Kevin Robinson and Yash Katariya and Sebastian Riedel and Paige Bailey and Kefan Xiao and Nimesh Ghelani and Lora Aroyo and Ambrose Slone and Neil Houlsby and Xuehan Xiong and Zhen Yang and Elena Gribovskaya and Jonas Adler and Mateo Wirth and Lisa Lee and Music Li and Thais Kagohara and Jay Pavagadhi and Sophie Bridgers and Anna Bortsova and Sanjay Ghemawat and Zafarali Ahmed and Tianqi Liu and Richard Powell and Vijay Bolina and Mariko Iinuma and Polina Zablotskaia and James Besley and Da-Woon Chung and Timothy Dozat and Ramona Comanescu and Xiance Si and Jeremy Greer and Guolong Su and Martin Polacek and Raphaël Lopez Kaufman and Simon Tokumine and Hexiang Hu and Elena Buchatskaya and Yingjie Miao and Mohamed Elhawaty and Aditya Siddhant and Nenad Tomasev and Jinwei Xing and Christina Greer and Helen Miller and Shereen Ashraf and Aurko Roy and Zizhao Zhang and Ada Ma and Angelos Filos and Milos Besta and Rory Blevins and Ted Klimenko and Chih-Kuan Yeh and Soravit Changpinyo and Jiaqi Mu and Oscar Chang and Mantas Pajarskas and Carrie Muir and Vered Cohen and Charline Le Lan and Krishna Haridasan and Amit Marathe and Steven Hansen and Sholto Douglas and Rajkumar Samuel and Mingqiu Wang and Sophia Austin and Chang Lan and Jiepu Jiang and Justin Chiu and Jaime Alonso Lorenzo and Lars Lowe Sjösund and Sébastien Cevey and Zach Gleicher and Thi Avrahami and Anudhyan Boral and Hansa Srinivasan and Vittorio Selo and Rhys May and Konstantinos Aisopos and Léonard Hussenot and Livio Baldini Soares and Kate Baumli and Michael B. Chang and Adrià Recasens and Ben Caine and Alexander Pritzel and Filip Pavetic and Fabio Pardo and Anita Gergely and Justin Frye and Vinay Ramasesh and Dan Horgan and Kartikeya Badola and Nora Kassner and Subhrajit Roy and Ethan Dyer and Víctor Campos Campos and Alex Tomala and Yunhao Tang and Dalia El Badawy and Elspeth White and Basil Mustafa and Oran Lang and Abhishek Jindal and Sharad Vikram and Zhitao Gong and Sergi Caelles and Ross Hemsley and Gregory Thornton and Fangxiaoyu Feng and Wojciech Stokowiec and Ce Zheng and Phoebe Thacker and Çağlar Ünlü and Zhishuai Zhang and Mohammad Saleh and James Svensson and Max Bileschi and Piyush Patil and Ankesh Anand and Roman Ring and Katerina Tsihlas and Arpi Vezer and Marco Selvi and Toby Shevlane and Mikel Rodriguez and Tom Kwiatkowski and Samira Daruki and Keran Rong and Allan Dafoe and Nicholas FitzGerald and Keren Gu-Lemberg and Mina Khan and Lisa Anne Hendricks and Marie Pellat and Vladimir Feinberg and James Cobon-Kerr and Tara Sainath and Maribeth Rauh and Sayed Hadi Hashemi and Richard Ives and Yana Hasson and Eric Noland and Yuan Cao and Nathan Byrd and Le Hou and Qingze Wang and Thibault Sottiaux and Michela Paganini and Jean-Baptiste Lespiau and Alexandre Moufarek and Samer Hassan and Kaushik Shivakumar and Joost van Amersfoort and Amol Mandhane and Pratik Joshi and Anirudh Goyal and Matthew Tung and Andrew Brock and Hannah Sheahan and Vedant Misra and Cheng Li and Nemanja Rakićević and Mostafa Dehghani and Fangyu Liu and Sid Mittal and Junhyuk Oh and Seb Noury and Eren Sezener and Fantine Huot and Matthew Lamm and Nicola De Cao and Charlie Chen and Sidharth Mudgal and Romina Stella and Kevin Brooks and Gautam Vasudevan and Chenxi Liu and Mainak Chain and Nivedita Melinkeri and Aaron Cohen and Venus Wang and Kristie Seymore and Sergey Zubkov and Rahul Goel and Summer Yue and Sai Krishnakumaran and Brian Albert and Nate Hurley and Motoki Sano and Anhad Mohananey and Jonah Joughin and Egor Filonov and Tomasz Kępa and Yomna Eldawy and Jiawern Lim and Rahul Rishi and Shirin Badiezadegan and Taylor Bos and Jerry Chang and Sanil Jain and Sri Gayatri Sundara Padmanabhan and Subha Puttagunta and Kalpesh Krishna and Leslie Baker and Norbert Kalb and Vamsi Bedapudi and Adam Kurzrok and Shuntong Lei and Anthony Yu and Oren Litvin and Xiang Zhou and Zhichun Wu and Sam Sobell and Andrea Siciliano and Alan Papir and Robby Neale and Jonas Bragagnolo and Tej Toor and Tina Chen and Valentin Anklin and Feiran Wang and Richie Feng and Milad Gholami and Kevin Ling and Lijuan Liu and Jules Walter and Hamid Moghaddam and Arun Kishore and Jakub Adamek and Tyler Mercado and Jonathan Mallinson and Siddhinita Wandekar and Stephen Cagle and Eran Ofek and Guillermo Garrido and Clemens Lombriser and Maksim Mukha and Botu Sun and Hafeezul Rahman Mohammad and Josip Matak and Yadi Qian and Vikas Peswani and Pawel Janus and Quan Yuan and Leif Schelin and Oana David and Ankur Garg and Yifan He and Oleksii Duzhyi and Anton Älgmyr and Timothée Lottaz and Qi Li and Vikas Yadav and Luyao Xu and Alex Chinien and Rakesh Shivanna and Aleksandr Chuklin and Josie Li and Carrie Spadine and Travis Wolfe and Kareem Mohamed and Subhabrata Das and Zihang Dai and Kyle He and Daniel von Dincklage and Shyam Upadhyay and Akanksha Maurya and Luyan Chi and Sebastian Krause and Khalid Salama and Pam G Rabinovitch and Pavan Kumar Reddy M and Aarush Selvan and Mikhail Dektiarev and Golnaz Ghiasi and Erdem Guven and Himanshu Gupta and Boyi Liu and Deepak Sharma and Idan Heimlich Shtacher and Shachi Paul and Oscar Akerlund and François-Xavier Aubet and Terry Huang and Chen Zhu and Eric Zhu and Elico Teixeira and Matthew Fritze and Francesco Bertolini and Liana-Eleonora Marinescu and Martin Bölle and Dominik Paulus and Khyatti Gupta and Tejasi Latkar and Max Chang and Jason Sanders and Roopa Wilson and Xuewei Wu and Yi-Xuan Tan and Lam Nguyen Thiet and Tulsee Doshi and Sid Lall and Swaroop Mishra and Wanming Chen and Thang Luong and Seth Benjamin and Jasmine Lee and Ewa Andrejczuk and Dominik Rabiej and Vipul Ranjan and Krzysztof Styrc and Pengcheng Yin and Jon Simon and Malcolm Rose Harriott and Mudit Bansal and Alexei Robsky and Geoff Bacon and David Greene and Daniil Mirylenka and Chen Zhou and Obaid Sarvana and Abhimanyu Goyal and Samuel Andermatt and Patrick Siegler and Ben Horn and Assaf Israel and Francesco Pongetti and Chih-Wei "Louis" Chen and Marco Selvatici and Pedro Silva and Kathie Wang and Jackson Tolins and Kelvin Guu and Roey Yogev and Xiaochen Cai and Alessandro Agostini and Maulik Shah and Hung Nguyen and Noah Ó Donnaile and Sébastien Pereira and Linda Friso and Adam Stambler and Adam Kurzrok and Chenkai Kuang and Yan Romanikhin and Mark Geller and ZJ Yan and Kane Jang and Cheng-Chun Lee and Wojciech Fica and Eric Malmi and Qijun Tan and Dan Banica and Daniel Balle and Ryan Pham and Yanping Huang and Diana Avram and Hongzhi Shi and Jasjot Singh and Chris Hidey and Niharika Ahuja and Pranab Saxena and Dan Dooley and Srividya Pranavi Potharaju and Eileen O'Neill and Anand Gokulchandran and Ryan Foley and Kai Zhao and Mike Dusenberry and Yuan Liu and Pulkit Mehta and Ragha Kotikalapudi and Chalence Safranek-Shrader and Andrew Goodman and Joshua Kessinger and Eran Globen and Prateek Kolhar and Chris Gorgolewski and Ali Ibrahim and Yang Song and Ali Eichenbaum and Thomas Brovelli and Sahitya Potluri and Preethi Lahoti and Cip Baetu and Ali Ghorbani and Charles Chen and Andy Crawford and Shalini Pal and Mukund Sridhar and Petru Gurita and Asier Mujika and Igor Petrovski and Pierre-Louis Cedoz and Chenmei Li and Shiyuan Chen and Niccolò Dal Santo and Siddharth Goyal and Jitesh Punjabi and Karthik Kappaganthu and Chester Kwak and Pallavi LV and Sarmishta Velury and Himadri Choudhury and Jamie Hall and Premal Shah and Ricardo Figueira and Matt Thomas and Minjie Lu and Ting Zhou and Chintu Kumar and Thomas Jurdi and Sharat Chikkerur and Yenai Ma and Adams Yu and Soo Kwak and Victor Ähdel and Sujeevan Rajayogam and Travis Choma and Fei Liu and Aditya Barua and Colin Ji and Ji Ho Park and Vincent Hellendoorn and Alex Bailey and Taylan Bilal and Huanjie Zhou and Mehrdad Khatir and Charles Sutton and Wojciech Rzadkowski and Fiona Macintosh and Roopali Vij and Konstantin Shagin and Paul Medina and Chen Liang and Jinjing Zhou and Pararth Shah and Yingying Bi and Attila Dankovics and Shipra Banga and Sabine Lehmann and Marissa Bredesen and Zifan Lin and John Eric Hoffmann and Jonathan Lai and Raynald Chung and Kai Yang and Nihal Balani and Arthur Bražinskas and Andrei Sozanschi and Matthew Hayes and Héctor Fernández Alcalde and Peter Makarov and Will Chen and Antonio Stella and Liselotte Snijders and Michael Mandl and Ante Kärrman and Paweł Nowak and Xinyi Wu and Alex Dyck and Krishnan Vaidyanathan and Raghavender R and Jessica Mallet and Mitch Rudominer and Eric Johnston and Sushil Mittal and Akhil Udathu and Janara Christensen and Vishal Verma and Zach Irving and Andreas Santucci and Gamaleldin Elsayed and Elnaz Davoodi and Marin Georgiev and Ian Tenney and Nan Hua and Geoffrey Cideron and Edouard Leurent and Mahmoud Alnahlawi and Ionut Georgescu and Nan Wei and Ivy Zheng and Dylan Scandinaro and Heinrich Jiang and Jasper Snoek and Mukund Sundararajan and Xuezhi Wang and Zack Ontiveros and Itay Karo and Jeremy Cole and Vinu Rajashekhar and Lara Tumeh and Eyal Ben-David and Rishub Jain and Jonathan Uesato and Romina Datta and Oskar Bunyan and Shimu Wu and John Zhang and Piotr Stanczyk and Ye Zhang and David Steiner and Subhajit Naskar and Michael Azzam and Matthew Johnson and Adam Paszke and Chung-Cheng Chiu and Jaume Sanchez Elias and Afroz Mohiuddin and Faizan Muhammad and Jin Miao and Andrew Lee and Nino Vieillard and Jane Park and Jiageng Zhang and Jeff Stanway and Drew Garmon and Abhijit Karmarkar and Zhe Dong and Jong Lee and Aviral Kumar and Luowei Zhou and Jonathan Evens and William Isaac and Geoffrey Irving and Edward Loper and Michael Fink and Isha Arkatkar and Nanxin Chen and Izhak Shafran and Ivan Petrychenko and Zhe Chen and Johnson Jia and Anselm Levskaya and Zhenkai Zhu and Peter Grabowski and Yu Mao and Alberto Magni and Kaisheng Yao and Javier Snaider and Norman Casagrande and Evan Palmer and Paul Suganthan and Alfonso Castaño and Irene Giannoumis and Wooyeol Kim and Mikołaj Rybiński and Ashwin Sreevatsa and Jennifer Prendki and David Soergel and Adrian Goedeckemeyer and Willi Gierke and Mohsen Jafari and Meenu Gaba and Jeremy Wiesner and Diana Gage Wright and Yawen Wei and Harsha Vashisht and Yana Kulizhskaya and Jay Hoover and Maigo Le and Lu Li and Chimezie Iwuanyanwu and Lu Liu and Kevin Ramirez and Andrey Khorlin and Albert Cui and Tian LIN and Marcus Wu and Ricardo Aguilar and Keith Pallo and Abhishek Chakladar and Ginger Perng and Elena Allica Abellan and Mingyang Zhang and Ishita Dasgupta and Nate Kushman and Ivo Penchev and Alena Repina and Xihui Wu and Tom van der Weide and Priya Ponnapalli and Caroline Kaplan and Jiri Simsa and Shuangfeng Li and Olivier Dousse and Fan Yang and Jeff Piper and Nathan Ie and Rama Pasumarthi and Nathan Lintz and Anitha Vijayakumar and Daniel Andor and Pedro Valenzuela and Minnie Lui and Cosmin Paduraru and Daiyi Peng and Katherine Lee and Shuyuan Zhang and Somer Greene and Duc Dung Nguyen and Paula Kurylowicz and Cassidy Hardin and Lucas Dixon and Lili Janzer and Kiam Choo and Ziqiang Feng and Biao Zhang and Achintya Singhal and Dayou Du and Dan McKinnon and Natasha Antropova and Tolga Bolukbasi and Orgad Keller and David Reid and Daniel Finchelstein and Maria Abi Raad and Remi Crocker and Peter Hawkins and Robert Dadashi and Colin Gaffney and Ken Franko and Anna Bulanova and Rémi Leblond and Shirley Chung and Harry Askham and Luis C. Cobo and Kelvin Xu and Felix Fischer and Jun Xu and Christina Sorokin and Chris Alberti and Chu-Cheng Lin and Colin Evans and Alek Dimitriev and Hannah Forbes and Dylan Banarse and Zora Tung and Mark Omernick and Colton Bishop and Rachel Sterneck and Rohan Jain and Jiawei Xia and Ehsan Amid and Francesco Piccinno and Xingyu Wang and Praseem Banzal and Daniel J. Mankowitz and Alex Polozov and Victoria Krakovna and Sasha Brown and MohammadHossein Bateni and Dennis Duan and Vlad Firoiu and Meghana Thotakuri and Tom Natan and Matthieu Geist and Ser tan Girgin and Hui Li and Jiayu Ye and Ofir Roval and Reiko Tojo and Michael Kwong and James Lee-Thorp and Christopher Yew and Danila Sinopalnikov and Sabela Ramos and John Mellor and Abhishek Sharma and Kathy Wu and David Miller and Nicolas Sonnerat and Denis Vnukov and Rory Greig and Jennifer Beattie and Emily Caveness and Libin Bai and Julian Eisenschlos and Alex Korchemniy and Tomy Tsai and Mimi Jasarevic and Weize Kong and Phuong Dao and Zeyu Zheng and Frederick Liu and Fan Yang and Rui Zhu and Tian Huey Teh and Jason Sanmiya and Evgeny Gladchenko and Nejc Trdin and Daniel Toyama and Evan Rosen and Sasan Tavakkol and Linting Xue and Chen Elkind and Oliver Woodman and John Carpenter and George Papamakarios and Rupert Kemp and Sushant Kafle and Tanya Grunina and Rishika Sinha and Alice Talbert and Diane Wu and Denese Owusu-Afriyie and Cosmo Du and Chloe Thornton and Jordi Pont-Tuset and Pradyumna Narayana and Jing Li and Saaber Fatehi and John Wieting and Omar Ajmeri and Benigno Uria and Yeongil Ko and Laura Knight and Amélie Héliou and Ning Niu and Shane Gu and Chenxi Pang and Yeqing Li and Nir Levine and Ariel Stolovich and Rebeca Santamaria-Fernandez and Sonam Goenka and Wenny Yustalim and Robin Strudel and Ali Elqursh and Charlie Deck and Hyo Lee and Zonglin Li and Kyle Levin and Raphael Hoffmann and Dan Holtmann-Rice and Olivier Bachem and Sho Arora and Christy Koh and Soheil Hassas Yeganeh and Siim Põder and Mukarram Tariq and Yanhua Sun and Lucian Ionita and Mojtaba Seyedhosseini and Pouya Tafti and Zhiyu Liu and Anmol Gulati and Jasmine Liu and Xinyu Ye and Bart Chrzaszcz and Lily Wang and Nikhil Sethi and Tianrun Li and Ben Brown and Shreya Singh and Wei Fan and Aaron Parisi and Joe Stanton and Vinod Koverkathu and Christopher A. Choquette-Choo and Yunjie Li and TJ Lu and Abe Ittycheriah and Prakash Shroff and Mani Varadarajan and Sanaz Bahargam and Rob Willoughby and David Gaddy and Guillaume Desjardins and Marco Cornero and Brona Robenek and Bhavishya Mittal and Ben Albrecht and Ashish Shenoy and Fedor Moiseev and Henrik Jacobsson and Alireza Ghaffarkhah and Morgane Rivière and Alanna Walton and Clément Crepy and Alicia Parrish and Zongwei Zhou and Clement Farabet and Carey Radebaugh and Praveen Srinivasan and Claudia van der Salm and Andreas Fidjeland and Salvatore Scellato and Eri Latorre-Chimoto and Hanna Klimczak-Plucińska and David Bridson and Dario de Cesare and Tom Hudson and Piermaria Mendolicchio and Lexi Walker and Alex Morris and Matthew Mauger and Alexey Guseynov and Alison Reid and Seth Odoom and Lucia Loher and Victor Cotruta and Madhavi Yenugula and Dominik Grewe and Anastasia Petrushkina and Tom Duerig and Antonio Sanchez and Steve Yadlowsky and Amy Shen and Amir Globerson and Lynette Webb and Sahil Dua and Dong Li and Surya Bhupatiraju and Dan Hurt and Haroon Qureshi and Ananth Agarwal and Tomer Shani and Matan Eyal and Anuj Khare and Shreyas Rammohan Belle and Lei Wang and Chetan Tekur and Mihir Sanjay Kale and Jinliang Wei and Ruoxin Sang and Brennan Saeta and Tyler Liechty and Yi Sun and Yao Zhao and Stephan Lee and Pandu Nayak and Doug Fritz and Manish Reddy Vuyyuru and John Aslanides and Nidhi Vyas and Martin Wicke and Xiao Ma and Evgenii Eltyshev and Nina Martin and Hardie Cate and James Manyika and Keyvan Amiri and Yelin Kim and Xi Xiong and Kai Kang and Florian Luisier and Nilesh Tripuraneni and David Madras and Mandy Guo and Austin Waters and Oliver Wang and Joshua Ainslie and Jason Baldridge and Han Zhang and Garima Pruthi and Jakob Bauer and Feng Yang and Riham Mansour and Jason Gelman and Yang Xu and George Polovets and Ji Liu and Honglong Cai and Warren Chen and XiangHai Sheng and Emily Xue and Sherjil Ozair and Christof Angermueller and Xiaowei Li and Anoop Sinha and Weiren Wang and Julia Wiesinger and Emmanouil Koukoumidis and Yuan Tian and Anand Iyer and Madhu Gurumurthy and Mark Goldenson and Parashar Shah and MK Blake and Hongkun Yu and Anthony Urbanowicz and Jennimaria Palomaki and Chrisantha Fernando and Ken Durden and Harsh Mehta and Nikola Momchev and Elahe Rahimtoroghi and Maria Georgaki and Amit Raul and Sebastian Ruder and Morgan Redshaw and Jinhyuk Lee and Denny Zhou and Komal Jalan and Dinghua Li and Blake Hechtman and Parker Schuh and Milad Nasr and Kieran Milan and Vladimir Mikulik and Juliana Franco and Tim Green and Nam Nguyen and Joe Kelley and Aroma Mahendru and Andrea Hu and Joshua Howland and Ben Vargas and Jeffrey Hui and Kshitij Bansal and Vikram Rao and Rakesh Ghiya and Emma Wang and Ke Ye and Jean Michel Sarr and Melanie Moranski Preston and Madeleine Elish and Steve Li and Aakash Kaku and Jigar Gupta and Ice Pasupat and Da-Cheng Juan and Milan Someswar and Tejvi M. and Xinyun Chen and Aida Amini and Alex Fabrikant and Eric Chu and Xuanyi Dong and Amruta Muthal and Senaka Buthpitiya and Sarthak Jauhari and Nan Hua and Urvashi Khandelwal and Ayal Hitron and Jie Ren and Larissa Rinaldi and Shahar Drath and Avigail Dabush and Nan-Jiang Jiang and Harshal Godhia and Uli Sachs and Anthony Chen and Yicheng Fan and Hagai Taitelbaum and Hila Noga and Zhuyun Dai and James Wang and Chen Liang and Jenny Hamer and Chun-Sung Ferng and Chenel Elkind and Aviel Atias and Paulina Lee and Vít Listík and Mathias Carlen and Jan van de Kerkhof and Marcin Pikus and Krunoslav Zaher and Paul Müller and Sasha Zykova and Richard Stefanec and Vitaly Gatsko and Christoph Hirnschall and Ashwin Sethi and Xingyu Federico Xu and Chetan Ahuja and Beth Tsai and Anca Stefanoiu and Bo Feng and Keshav Dhandhania and Manish Katyal and Akshay Gupta and Atharva Parulekar and Divya Pitta and Jing Zhao and Vivaan Bhatia and Yashodha Bhavnani and Omar Alhadlaq and Xiaolin Li and Peter Danenberg and Dennis Tu and Alex Pine and Vera Filippova and Abhipso Ghosh and Ben Limonchik and Bhargava Urala and Chaitanya Krishna Lanka and Derik Clive and Yi Sun and Edward Li and Hao Wu and Kevin Hongtongsak and Ianna Li and Kalind Thakkar and Kuanysh Omarov and Kushal Majmundar and Michael Alverson and Michael Kucharski and Mohak Patel and Mudit Jain and Maksim Zabelin and Paolo Pelagatti and Rohan Kohli and Saurabh Kumar and Joseph Kim and Swetha Sankar and Vineet Shah and Lakshmi Ramachandruni and Xiangkai Zeng and Ben Bariach and Laura Weidinger and Tu Vu and Alek Andreev and Antoine He and Kevin Hui and Sheleem Kashem and Amar Subramanya and Sissie Hsiao and Demis Hassabis and Koray Kavukcuoglu and Adam Sadovsky and Quoc Le and Trevor Strohman and Yonghui Wu and Slav Petrov and Jeffrey Dean and Oriol Vinyals},
      year={2025},
      eprint={2312.11805},
      archivePrefix={arXiv},
      primaryClass={cs.CL},
      url={https://arxiv.org/abs/2312.11805}, 
}

@misc{gpt,
      title={{OpenAI} {GPT}-5 System Card}, 
      author={Aaditya Singh and Adam Fry and Adam Perelman and Adam Tart and Adi Ganesh and Ahmed El-Kishky and Aidan McLaughlin and Aiden Low and AJ Ostrow and Akhila Ananthram and Akshay Nathan and Alan Luo and Alec Helyar and Aleksander Madry and Aleksandr Efremov and Aleksandra Spyra and Alex Baker-Whitcomb and Alex Beutel and Alex Karpenko and Alex Makelov and Alex Neitz and Alex Wei and Alexandra Barr and Alexandre Kirchmeyer and Alexey Ivanov and Alexi Christakis and Alistair Gillespie and Allison Tam and Ally Bennett and Alvin Wan and Alyssa Huang and Amy McDonald Sandjideh and Amy Yang and Ananya Kumar and Andre Saraiva and Andrea Vallone and Andrei Gheorghe and Andres Garcia Garcia and Andrew Braunstein and Andrew Liu and Andrew Schmidt and Andrey Mereskin and Andrey Mishchenko and Andy Applebaum and Andy Rogerson and Ann Rajan and Annie Wei and Anoop Kotha and Anubha Srivastava and Anushree Agrawal and Arun Vijayvergiya and Ashley Tyra and Ashvin Nair and Avi Nayak and Ben Eggers and Bessie Ji and Beth Hoover and Bill Chen and Blair Chen and Boaz Barak and Borys Minaiev and Botao Hao and Bowen Baker and Brad Lightcap and Brandon McKinzie and Brandon Wang and Brendan Quinn and Brian Fioca and Brian Hsu and Brian Yang and Brian Yu and Brian Zhang and Brittany Brenner and Callie Riggins Zetino and Cameron Raymond and Camillo Lugaresi and Carolina Paz and Cary Hudson and Cedric Whitney and Chak Li and Charles Chen and Charlotte Cole and Chelsea Voss and Chen Ding and Chen Shen and Chengdu Huang and Chris Colby and Chris Hallacy and Chris Koch and Chris Lu and Christina Kaplan and Christina Kim and CJ Minott-Henriques and Cliff Frey and Cody Yu and Coley Czarnecki and Colin Reid and Colin Wei and Cory Decareaux and Cristina Scheau and 385 others},
      year={2026},
      eprint={2601.03267},
      archivePrefix={arXiv},
      primaryClass={cs.CL},
      url={https://arxiv.org/abs/2601.03267}, 
}

@misc{fanar2,
      title={Fanar 2.0: {A}rabic {G}enerative {AI} {S}tack}, 
      author={{FANAR TEAM} and Ummar Abbas and Mohammad Shahmeer Ahmad and Minhaj Ahmad and Abdulaziz Al-Homaid and Anas Al-Nuaimi and Enes Altinisik and Ehsaneddin Asgari and Sanjay Chawla and Shammur Chowdhury and Fahim Dalvi and Kareem Darwish and Nadir Durrani and Mohamed Elfeky and Ahmed Elmagarmid and Mohamed Eltabakh and Asim Ersoy and Masoomali Fatehkia and Mohammed Qusay Hashim and Majd Hawasly and Mohamed Hefeeda and Mus'ab Husaini and Keivin Isufaj and Soon-Gyo Jung and Houssam Lachemat and Ji Kim Lucas and Abubakr Mohamed and Tasnim Mohiuddin and Basel Mousi and Hamdy Mubarak and Ahmad Musleh and Mourad Ouzzani and Amin Sadeghi and Husrev Taha Sencar and Mohammed Shinoy and Omar Sinan and Yifan Zhang},
      year={2026},
      eprint={2603.16397},
      archivePrefix={arXiv},
      primaryClass={cs.CL},
      url={https://arxiv.org/abs/2603.16397}, 
}

@misc{meta2025llama4,
  author = {{Meta}},
  title = {The {L}lama 4 herd: The beginning of a new era of natively multimodal {AI} innovation},
  year = {2025},
  url = {https://ai.meta.com/blog/llama-4-multimodal-intelligence/},
  note = {Accessed: 2025-04-29}
}

@misc{dozat2017deepbiaffineattentionneural,
      title={Deep {B}iaffine {A}ttention for {N}eural {D}ependency {P}arsing}, 
      author={Timothy Dozat and Christopher D. Manning},
      year={2017},
      eprint={1611.01734},
      archivePrefix={arXiv},
      primaryClass={cs.CL},
      url={https://arxiv.org/abs/1611.01734}, 
}

@inproceedings{kwon2023efficient,
  title={Efficient memory management for large language model serving with pagedattention},
  author={Kwon, Woosuk and Li, Zhuohan and Zhuang, Siyuan and Sheng, Ying and Zheng, Lianmin and Yu, Cody Hao and Gonzalez, Joseph and Zhang, Hao and Stoica, Ion},
  booktitle={Proceedings of the 29th symposium on operating systems principles},
  pages={611--626},
  year={2023}
}

@inproceedings{Obeid:2020:camel,
    title = "{CAM}e{L} Tools: An Open Source Python Toolkit for {A}rabic Natural Language Processing",
    author = "Obeid, Ossama  and
      Zalmout, Nasser  and
      Khalifa, Salam  and
      Taji, Dima  and
      Oudah, Mai  and
      Alhafni, Bashar  and
      Inoue, Go  and
      Eryani, Fadhl  and
      Erdmann, Alexander  and
      Habash, Nizar",
    editor = "Calzolari, Nicoletta  and
      B{\'e}chet, Fr{\'e}d{\'e}ric  and
      Blache, Philippe  and
      Choukri, Khalid  and
      Cieri, Christopher  and
      Declerck, Thierry  and
      Goggi, Sara  and
      Isahara, Hitoshi  and
      Maegaard, Bente  and
      Mariani, Joseph  and
      Mazo, H{\'e}l{\`e}ne  and
      Moreno, Asuncion  and
      Odijk, Jan  and
      Piperidis, Stelios",
    booktitle = "Proceedings of the Twelfth Language Resources and Evaluation Conference",
    month = may,
    year = "2020",
    address = "Marseille, France",

    url = "https://aclanthology.org/2020.lrec-1.868",
    pages = "7022--7032",
    language = "English",
    ISBN = "979-10-95546-34-4",
}

@inproceedings{Gerdes:2021:SUD,
  TITLE = {{Starting a new treebank? Go SUD! Theoretical and practical benefits of the Surface-Syntactic distributional approach}},
  AUTHOR = {Gerdes, Kim and Guillaume, Bruno and Kahane, Sylvain and Perrier, Guy},
  URL = {https://hal.inria.fr/hal-03509136},
  BOOKTITLE = {{Sixth International Conference on Dependency Linguistics (Depling, SyntaxFest 2021)}},
  ADDRESS = {Sofia, Bulgaria},
  YEAR = {2021},

}

@inproceedings{CamelTB,
    title = "Camel Treebank: An Open Multi-genre {A}rabic Dependency Treebank",
    author = "Habash, Nizar  and
      AbuOdeh, Muhammed  and
      Taji, Dima  and
      Faraj, Reem  and
      El Gizuli, Jamila  and
      Kallas, Omar",
    editor = "Calzolari, Nicoletta  and
      B{\'e}chet, Fr{\'e}d{\'e}ric  and
      Blache, Philippe  and
      Choukri, Khalid  and
      Cieri, Christopher  and
      Declerck, Thierry  and
      Goggi, Sara  and
      Isahara, Hitoshi  and
      Maegaard, Bente  and
      Mariani, Joseph  and
      Mazo, H{\'e}l{\`e}ne  and
      Odijk, Jan  and
      Piperidis, Stelios",
    booktitle = "Proceedings of the Thirteenth Language Resources and Evaluation Conference",
    month = jun,
    year = "2022",
    address = "Marseille, France",

    url = "https://aclanthology.org/2022.lrec-1.286",
    pages = "2672--2681"
}

@article{adel2026arabic,
  title={Arabic {M}orphosyntactic {T}agging and {D}ependency {P}arsing with {L}arge {L}anguage {M}odels},
  author={Adel, Mohamed and Alhafni, Bashar and Habash, Nizar},
  journal={arXiv preprint arXiv:2603.16718},
  year={2026}
}

@article{de2021universal,
  title={Universal dependencies},
  author={De Marneffe, Marie-Catherine and Manning, Christopher D and Nivre, Joakim and Zeman, Daniel},
  journal={Computational linguistics},
  volume={47},
  number={2},
  pages={255--308},
  year={2021}
}

@inproceedings{faroukzakariaelshabrawy-etal-2026-parsing,
  title = {Parsing {A}rabic {D}ialects {R}evisited: {N}ew {B}enchmarks, {M}odels, and {I}nsights },
  author = {Elshabrawy, Ahmed and Inoue, Go and AbuOdeh, Muhammed and Habash, Nizar},
  booktitle = {The 7th Workshop on Open-Source Arabic Corpora and Processing Tools (OSACT7) with 5 Shared Tasks},
  month = {May},
  year = {2026},
  pages = {94--105},
  address = {Palma, Mallorca, Spain},
  publisher = {European Language Resources Association (ELRA)},
  editor = {Al-Khalifa, Hend and El-Haj, Mo and Ezzini, Saad},
  doi = {10.63317/3eyeu3k726ab}
}

@inproceedings{akhundjanova-etal-2025-parallel,
    title = "Parallel {U}niversal {D}ependencies {T}reebanks for {T}urkic {L}anguages",
    author = "Akhundjanova, Arofat  and
      Akkurt, Furkan  and
      Chontaeva, Bermet  and
      Eslami, Soudabeh  and
      Coltekin, Cagri",
    editor = {Bouma, Gosse  and
      {\c{C}}{\"o}ltekin, {\c{C}}a{\u{g}}r{\i}},
    booktitle = "Proceedings of the Eighth Workshop on Universal Dependencies (UDW, SyntaxFest 2025)",
    month = aug,
    year = "2025",
    address = "Ljubljana, Slovenia",
    publisher = "Association for Computational Linguistics",
    url = "https://aclanthology.org/2025.udw-1.14/",
    pages = "129--136",
    ISBN = "979-8-89176-292-3",
}

@book{Jurafsky:2009:SLP:1214993,
 author = {Jurafsky, Daniel and Martin, James H.},
 title = {Speech and Language Processing (2Nd Edition)},
 year = {2009},
 isbn = {0131873210},
 publisher = {Prentice-Hall, Inc.},
 address = {Upper Saddle River, NJ, USA},
}

@inproceedings{li-etal-2025-generation,
    title = "From Generation to Judgment: Opportunities and Challenges of {LLM}-as-a-judge",
    author = "Li, Dawei  and
      Jiang, Bohan  and
      Huang, Liangjie  and
      Beigi, Alimohammad  and
      Zhao, Chengshuai  and
      Tan, Zhen  and
      Bhattacharjee, Amrita  and
      Jiang, Yuxuan  and
      Chen, Canyu  and
      Wu, Tianhao  and
      Shu, Kai  and
      Cheng, Lu  and
      Liu, Huan",
    editor = "Christodoulopoulos, Christos  and
      Chakraborty, Tanmoy  and
      Rose, Carolyn  and
      Peng, Violet",
    booktitle = "Proceedings of the 2025 Conference on Empirical Methods in Natural Language Processing",
    month = nov,
    year = "2025",
    address = "Suzhou, China",
    publisher = "Association for Computational Linguistics",
    url = "https://aclanthology.org/2025.emnlp-main.138/",
    doi = "10.18653/v1/2025.emnlp-main.138",
    pages = "2757--2791",
    ISBN = "979-8-89176-332-6"
}

@phdthesis{kulmizev2023search,
  title = {The Search for Syntax: Investigating the Syntactic Knowledge of Neural Language Models Through the Lens of Dependency Parsing},
  author = {Kulmizev, Artur},
  school = {Uppsala University},
  year = {2023},
  month = {09},
  publisher = {Acta Universitatis Upsaliensis},
  address = {Uppsala, Sweden},
  pages = {101},
  series = {Studia Linguistica Upsaliensia; 30},
  isbn = {978-91-513-1850-9},
  language = {en},
  url = {https://uu.diva-portal.org/smash/get/diva2:1784732/FULLTEXT01.pdf}
}

@inproceedings{kanayama-etal-2024,
    title = "Incorporating Syntax and Lexical Knowledge to Multilingual Sentiment Classification on Large Language Models",
    author = "Kanayama, Hiroshi  and
      Zhao, Yang  and
      Iwamoto, Ran  and
      Ohko, Takuya",
    editor = "Ku, Lun-Wei  and
      Martins, Andre  and
      Srikumar, Vivek",
    booktitle = "Findings of the Association for Computational Linguistics: ACL 2024",
    month = aug,
    year = "2024",
    address = "Bangkok, Thailand",
    publisher = "Association for Computational Linguistics",
    url = "https://aclanthology.org/2024.findings-acl.286/",
    doi = "10.18653/v1/2024.findings-acl.286",
    pages = "4810--4817"
}

@article{Azzoug_2010, title={Modern {S}tandard {A}rabic as a {S}econd {L}anguage for {I}ts {O}wn {S}peakers}, volume={16}, url={https://asjp.cerist.dz/en/article/186311}, number={1}, journal={AL-Lisaniyyat}, author={Azzoug, Omar}, year={2010}, month=jun, pages={1–22}, language={en} }

@misc{openai2026gpt56,
  author       = {{OpenAI}},
  title        = {{GPT-5.6 System Card}},
  year         = {2026},
  month        = jul,
  day          = {9},
  howpublished = {\url{https://deploymentsafety.openai.com/gpt-5-6/introduction}},
  note         = {Accessed: 2026-08-03}
}

@software{camellab_arabic_c2d,
  author  = {{CAMeL Lab}},
  title   = {{Arabic-C2D}: Arabic Constituency to Dependency Mapper and
             {PATB}-to-{CATiB} Map},
  year    = {2019},
  month   = apr,
  day     = {4},
  url     = {https://github.com/CAMeL-Lab/Arabic-C2D},
  urldate = {2026-08-03}
}

@misc{muller2020multilinguallanguagemodelstransfer,
      title={Can Multilingual Language Models Transfer to an Unseen Dialect? A Case Study on {N}orth {A}frican {A}rabizi}, 
      author={Benjamin Muller and Benoit Sagot and Djamé Seddah},
      year={2020},
      eprint={2005.00318},
      archivePrefix={arXiv},
      primaryClass={cs.CL},
      url={https://arxiv.org/abs/2005.00318}, 
}
